\pdfoutput=1

\documentclass[11pt]{article}

\usepackage[preprint]{acl}

\usepackage{times}
\usepackage{latexsym}

\usepackage[T1]{fontenc}

\usepackage[utf8]{inputenc}

\usepackage{microtype}

\usepackage{inconsolata}

\usepackage{graphicx}

\usepackage{listings}
\usepackage{enumitem}

\usepackage{booktabs}
\usepackage{xcolor}
\usepackage{amsmath}
\usepackage{url}
\usepackage{caption}
\usepackage{subcaption}
\usepackage{multirow}
\usepackage[multiple]{footmisc}
\usepackage{soul}

\title{``My life is miserable, have to sign 500 autographs everyday'': Exposing Humblebragging, the Brags in Disguise}

\author{
 \textbf{Sharath Naganna*},
 \textbf{Saprativa Bhattacharjee*},
 \textbf{Biplab Banerjee},
 \textbf{Pushpak Bhattacharyya}
\\
 Indian Institute of Technology Bombay, Mumbai, India \\
\texttt{\{sharathhn, saprativa, pb\}@cse.iitb.ac.in, bbanerjee@iitb.ac.in}
}

\begin{document}
\maketitle
\def\thefootnote{$\ast$}\footnotetext{Equal contribution.}\def\thefootnote{\arabic{footnote}}

\begin{abstract}
\textit{Humblebragging} is a phenomenon in which individuals present self-promotional statements under the guise of modesty or complaints. For example, a statement like, ``\textit{Ugh, I can't believe I got promoted to lead the entire team. So stressful!}'', subtly highlights an achievement while pretending to be complaining. Detecting humblebragging is important for machines to better understand the nuances of human language, especially in tasks like sentiment analysis and intent recognition. However, this topic has not yet been studied in computational linguistics. For the first time, we introduce the task of automatically detecting humblebragging in text. We formalize the task by proposing a 4-tuple definition of humblebragging and evaluate machine learning, deep learning, and large language models (LLMs) on this task, comparing their performance with humans. We also create and release a dataset called HB-24, containing 3,340 humblebrags generated using GPT-4o. Our experiments show that detecting humblebragging is non-trivial, even for humans. Our best model achieves an F1-score of 0.88. This work lays the foundation for further exploration of this nuanced linguistic phenomenon and its integration into broader natural language understanding systems.
\end{abstract}

\section{Introduction}

\textit{Humblebragging} is a nuanced socio-linguistic phenomenon in which individuals subtly boast about their achievements, possessions, or qualities while disguising their self-promotion with expressions of complaint or modesty. The term was coined by American comedian Harris Wittels in 2010, who later published a book on the phenomenon titled \textit{Humblebrag: The Art of False Modesty} \cite{wittels2012humblebrag}. A few examples of humblebrags are provided in \autoref{tab:examples}.

\begin{figure}[ht]
    \centering
    \includegraphics[width=0.8\linewidth]{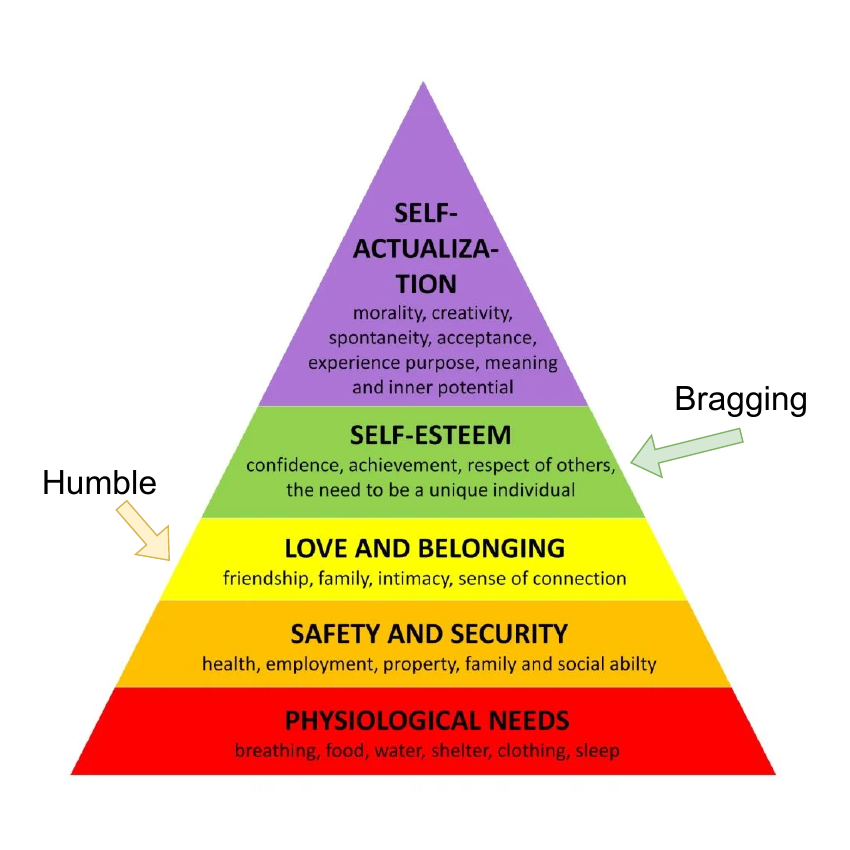}
    \caption{Why do people resort to humblebragging? The answer lies in the Maslow’s Hierarchy of Needs\protect\footnotemark. Humblebragging satisfies needs of belonging and self-esteem simultaneously.}
    \label{fig:maslows}
\end{figure}

\footnotetext{Image source: \url{https://www.simplypsychology.org/maslow.html}}

To address why people resort to humblebragging, \citet{trivedi2019bragging} analyzed its rising popularity, particularly on social media, from the dual perspectives of need theories (\textit{Maslow's Hierarchy of Needs}, \citealt{maslow1987maslow}; see \autoref{fig:maslows}) and the \textit{Hubris Hypothesis} \cite{hoorens2012hubris}. Humblebragging allows a person to satisfy the dual needs of belonging (level 3) through humility; and self-esteem (level 4) through bragging. Moreover, the hubris hypothesis states that addressees prefer implicit self-superiority claims over explicit ones. Also, straightforward bragging violates the maxims of modesty \cite{Leech1983PrinciplesOP} and self-denigration \cite{GU1990237}, as noted by \citet{ZUO2023165}. Humblebragging thus serves as a subtle way to convey positive information about oneself, often through expressions of modesty or complaint. Even though it is commonly observed on social media (Twitter, Reddit, and Instagram), many people also use it in real-life conversations \cite{wittels2012humblebrag, 2017-40996-001, trivedi2019bragging}. Although humblebragging has been recognized socially and culturally, it remains unexplored through the lens of computational linguistics. Evidence from search trends, social media behavior, and historical usage suggests that humblebragging is a persistent and recognizable phenomenon, even when not explicitly labeled. We provide supporting statistics and examples in \autoref{app: Prevalence}.

\begin{table*}[ht]
\centering
\resizebox{0.9\textwidth}{!}{%
\begin{tabular}{@{}l|l@{}}
\toprule
\multicolumn{1}{c}{Humblebrags} & Mask type \\ \midrule
{\color{blue}I can't believe} {\color{red}they'd give} {\color{blue}an idiot like me} {\color{red}a phd} {\color{blue}lol} & Modesty \\ \midrule
{\color{red}Being in demand means disappointing 95\% of people 95\% of the time.} {\color{blue}I have yet to learn how to overcome this.} & Modesty \\ \midrule
{\color{red}For the 3rd time in 3 years I've been asked to speak at Harvard,} {\color{blue}but I've yet to speak at my alma mater. What's a girl gotta do @MarquetteU?} & Complaint \\ \midrule
{\color{red}Will Twitter be available for me in Paris, milan, or the Maldives?} {\color{blue}I hope so bc it won't in hong Kong or Singapore} & Complaint \\ \bottomrule
\end{tabular}%
}
\caption{Examples of humblebrags. Each instance of a humblebrag consists of a brag masked by either complaint or modesty. The brags are in {\color{red}red} while the masks are in {\color{blue}blue}.}
\label{tab:examples}
\end{table*}

Humblebragging is closely related to other forms of figurative language, such as \textit{sarcasm} \cite{gibbs1986psycholinguistics} and \textit{irony} \cite{garmendia2018irony}, which also rely on \textit{verbal incongruity}, a contrast between what is said and what is meant. While irony typically contrasts expectation with reality and sarcasm adds a mocking tone, humblebragging uniquely conceals self-promotion within a modest or complaining remark. A detailed comparison of these differences is provided in \autoref{app: Differences}. Although extensive research exists on computational modeling of sarcasm \cite{bhattacharyya-joshi-2017-computational, riloff-etal-2013-sarcasm, cai-etal-2019-multi} and irony \cite{zeng-li-2022-survey, van-hee-etal-2018-semeval, barbieri-saggion-2014-modelling}, to the best of our knowledge, this work is the first to explore humblebragging in computational linguistics.

 Detecting humblebragging automatically is crucial for a nuanced understanding  and processing of human language. This is necessary for enhancing accuracy of tasks such as sentiment analysis \cite{zhang-etal-2024-sentiment}, intent recognition \cite{lamanov-etal-2022-template}, emotion recognition \cite{li-etal-2022-emocaps} and dialogue understanding \cite{gao-etal-2024-self}. In applications like social media monitoring and customer feedback analysis, it helps differentiate genuine complaints \cite{singh-etal-2023-reimagining} from brags disguised as complaints. Moreover, this capability is also valuable for researchers in the humanities and social sciences.

In this paper, we formally introduce the task of humblebragging detection and present a curated dataset that combines existing resources with synthetic data generated using GPT-4o. By addressing this previously unexplored area, our work bridges the gap between computational linguistics and other disciplines in the study of humblebragging. 

Our contributions are:

\begin{enumerate}
    \item Introduction of the task of automatic humblebragging detection from text to the computational linguistics community by proposing a 4-tuple definition of humblebragging for streamlining its processing (\autoref{Task Formulation} and \autoref{Methodology}).
    \item Benchmarking of various machine learning, deep learning, and state-of-the-art large language model techniques on the task of automatic humblebragging detection from text (\autoref{Experimental Setup} and \autoref{Results and Discussion}).
    \item Release of a new dataset named HB-24 on humblebragging detection, containing 3,340 humblebrags, to enable further research on the task (\autoref{Dataset}).
\end{enumerate}

\section{Related Work}
 \paragraph{In Psychology and Other Disciplines} Humblebragging has been extensively studied in psychology. \citet{2017-40996-001} examined its effects on audiences, showing that humblebragging is ubiquitous in daily interactions, with 70\% of humblebrags falling into the complaint-masked variety. \citet{trivedi2019bragging} explained the widespread use of humblebragging through the dual perspectives of need theories and the hubris hypothesis, and also provided a contextual framework for understanding it. Other notable studies include \citet{sezer2015psychology, vranka_becková_houdek_2017, Luo2020Humblebragging}. Beyond psychology, humblebragging has been explored in disciplines like tourism research \cite{chen2020bragging, yan2024bragging}, pragmatics \cite{lin2022also, ZUO2023165, han2024also}, and advertising \cite{paramita2021benefits}.

\paragraph{Sarcasm and Irony} Both \textit{sarcasm} and \textit{irony} have been extensively studied in computational linguistics over the past two decades. \citet{joshi-etal-2015-harnessing} demonstrated how incongruences can enhance sarcasm detection, while \citet{10.1145/3124420} categorized detection methods, benchmark datasets, and evaluation metrics. Recently, \citet{gole2023sarcasmdetectionopenaigptbased} explored the use of large language models for sarcasm detection. Beyond detection, \citet{joshi2017automaticidentificationsarcasmtarget} proposed a hybrid rule-based and statistical approach for identifying sarcasm targets, which was later complemented by transformer-based methods such as BERT, as demonstrated by \citet{parameswaran-etal-2021-berts}. For irony, \citet{zeng-li-2022-survey} provided a comprehensive survey on computational approaches. \citet{10.1007/978-3-319-19390-8_38} evaluated traditional machine learning models for irony detection using sentiment scores, while \citet{electronics12122673} introduced the \textit{Retrieval–Detection Method for Verbal Irony (RDVI)}, leveraging open-domain resources for enriched detection.

\paragraph{Bragging and Humility} A closely related area of research is the detection and processing of \textit{bragging} \cite{alfano2014bragging} and \textit{humility} \cite{snow1995humility} as standalone tasks. \citet{jin-etal-2022-automatic} introduced bragging classification to the computational linguistics community and released a public dataset, while \citet{jin-etal-2024-bragging} conducted a large-scale study of bragging behavior on Twitter. For humility, \citet{guo-etal-2024-computational} explored LLM-based techniques for measuring humility in social media posts. Additionally, \citet{danescu-niculescu-mizil-etal-2013-computational, firdaus-etal-2022-polise, srinivasan-choi-2022-tydip} have examined \textit{politeness} in computational contexts.

\paragraph{Synthetic Data Generation}
The language generation capabilities of LLMs have created opportunities for generating synthetic data. \citet{long-etal-2024-llms} provide a comprehensive survey of synthetic data generation, curation, and evaluation, while \citet{li-etal-2023-synthetic} explore the potential and limitations of using LLMs for this purpose. Synthetic data generated by language models has been applied to various text classification tasks \cite{chung-etal-2023-increasing, sahu-etal-2022-data, ye-etal-2022-zerogen, yoo-etal-2021-gpt3mix-leveraging}.

\paragraph{Classification as Generation} Finally, there has been a growing trend towards performing classification tasks by posing them as generation tasks. This approach to text classification is particularly relevant for leveraging decoder-based large language models in classification settings, where text generation mechanisms complement traditional methods. For instance, LLMs have been employed as zero-shot \cite{gretz-etal-2023-zero} and few-shot \cite{mirza-etal-2024-illuminer} text classifiers. Moreover, \citet{saunshi2021mathematicalexplorationlanguagemodels} provides mathematical insights into modeling classification tasks as text completion tasks.

\section{Formulation of the Humblebragging Definition}
\label{Task Formulation}
In this section, we define and derive our proposed framework for humblebragging.

\subsection{Formal Definition of Humblebragging}
\label{subsec:humblebragging}

We define humblebragging as a 4-tuple to systematically capture its key components and underlying structure:
\begin{equation}
HB = \langle B, BT, HM, MT \rangle
\end{equation}
where:
\begin{itemize}[noitemsep, topsep=0pt]
    \item \( B \): \textbf{Brag} – The segment of the text that explicitly conveys the act of bragging.
    \item \( BT \): \textbf{Brag Theme} – The overarching theme or specific category of the brag embedded within the statement. Categories are listed in \autoref{apx:categories}.
    \item \( HM \): \textbf{Humble Mask} – The segment of the text that adopts a modest or complaining tone to obscure or mitigate the act of bragging.
    \item \( MT \): \textbf{Mask Type} – Specifies whether the humble mask adopts a modest tone or a complaining approach.
\end{itemize}
\vspace{1em}

For instance, in the following statement: 
\begin{quote}
\textit{"Ugh, I can't believe I got promoted to lead the entire team. So stressful!"}
\end{quote}
\begin{itemize}[noitemsep, topsep=0pt]
    \item $\textbf{B}$: \textit{"I can't believe I got promoted to lead the entire team."};
    \item $\textbf{BT}$: Performance at work;
    \item $\textbf{HM}$: \textit{"Ugh,"} and \textit{"So stressful!"};
    \item $\textbf{MT}$: Complaint.
\end{itemize}



\subsection{Derivation of 4-tuple Definition}

Our 4-tuple definition is adapted from the 6-tuple framework of sarcasm \cite{article1}: \textbf{$\langle$Context (C), Utterance (u), Literal Proposition (p), Intended Proposition (p'), Speaker (S), and Hearer (H)$\rangle$}. In sarcasm, $p$ conveys a surface-level meaning that contrasts with $p'$, creating incongruity. In humblebragging, the same phenomenon is achieved through a \textbf{Brag (B)} and a \textbf{Humble Mask (HM)}, where HM corresponds to $p$, presenting a modest or complaining front, while B aligns with $p'$, subtly revealing the self-promotion. The Context (C) maps to the \textbf{Brag Theme (BT)}, which categorizes the nature of the brag (e.g., achievements, wealth, intelligence). Additionally, the \textbf{Mask Type (MT)} incorporates the classifications into modesty or complaint, further refining the nature of the humblebrag.  

Unlike sarcasm, where the Speaker (S) directs an utterance toward a Hearer (H) who must infer the intended meaning, humblebragging often lacks a specific hearer. It is frequently self-directed or broadcasted to a broad audience, making S and H unnecessary in this framework. This adaptation preserves the dual-layered meaning from sarcasm while formalizing humblebragging as a strategic blend of self-effacement and self-promotion.

\section{Dataset}
\label{Dataset}
As there were no existing datasets for the task of \textit{humblebragging detection}, we propose \textit{HB-24}\footnote{Available at \url{https://github.com/SharathHN/HB-24}}, a well-balanced collection of humblebrag and non-humblebrag texts, comprising both human-written and synthetic samples. Due to the limited availability of quality humblebrags, we leverage the capabilities of large language models to generate human-like examples, augmenting the existing data and enhancing the corpus for training classification models. The synthetic data is used for training, while the trained model's performance is evaluated on human-written samples. In other words, the training set is composed of synthetic humblebrags while the test set consists of human-written humblebrags. The non-humblebrags are all human-written. \autoref{tab:dataset} presents the dataset statistics.

\begin{table}[t]
\centering
\resizebox{0.8\columnwidth}{!}{%
\begin{tabular}{@{}llrr@{}}
\toprule
 &  & Humblebrag & Non-humblebrag \\ \midrule
\multicolumn{1}{c}{\multirow{4}{*}{Train}} & Samples & 3340 & 5431 \\
\multicolumn{1}{c}{} & Min \#words & 6 & 1 \\
\multicolumn{1}{c}{} & Max \#words & 47 & 68 \\
\multicolumn{1}{c}{} & Avg \#words & 15.98 & 16.41 \\ \midrule
\multirow{4}{*}{Test} & Samples & 558 & 576 \\
 & Min \#words & 1 & 6 \\
 & Max \#words & 70 & 47 \\
 & Avg \#words & 19.55 & 17.5 \\ \bottomrule
\end{tabular}%
}
\caption{Dataset statistics. Min, Max and Avg refer to minimum, maximum and average respectively.}
\label{tab:dataset}
\end{table}

\subsection{Human-Written Humblebrags}
\citet{wittels2012humblebrag} presents a curated collection of high-quality humblebrag texts, categorized into themes such as wealth, first-class travel, workplace achievements, celebrity status, and more. These tweets form the positive class within our test set.

\subsection{Synthetic Humblebrags}  
The humblebrags in the training set consist entirely of synthetic tweets generated using GPT-4o through zero-shot and few-shot prompting. The prompt template follows a format similar to that of \citet{li-etal-2023-synthetic}. In the following sections, we discuss the prompts used for generating synthetic humblebrags.

\subsubsection{Zero-Shot Prompts}  
In the zero-shot generation setup, we used two types of prompts. In the \textit{General Prompt}, we did not explicitly define humblebragging; instead, we asked the model to generate tweets that subtly mention various achievements. In the \textit{Prompt with Themes}, we provided a formal definition of humblebragging along with the themes (\autoref{apx:categories}) outlined in \citet{wittels2012humblebrag}. The prompts are provided in \autoref{app:data-prompts}.

\subsubsection{Few-Shot Prompts}
In the few-shot prompt setup, we modified the \textit{Prompt with Themes} to include a few examples from each theme. We experimented with varying numbers of examples, starting with one and going up to five, and observed that increasing the number of examples did not improve the generation quality. Consequently, we settled on three examples per prompt for generating samples with few-shot prompts.

\begin{table}[t]
\centering
\resizebox{0.5\columnwidth}{!}{%
\begin{tabular}{@{}lr@{}}
\toprule
Prompt Type              & \#Samples \\ \midrule
General Prompt           & 1100      \\
Prompt with Themes   & 1304      \\
Few-Shot with Themes & 936       \\
Total                    & 3340      \\ \bottomrule
\end{tabular}%
}
\caption{Prompt type and the number of samples.}
\label{tab:promt-categories}
\end{table}

\subsubsection{Post-Processing and Data Curation}  
After executing all three prompts, we generated a total of 11,000 synthetic samples containing humblebrags. Each sample was manually reviewed to assess its quality and relevance (see \autoref{app:Dataset Quality Assurance} for more details). From this pool, we filtered tweets from each prompt type, ensuring a balanced selection (see \autoref{tab:promt-categories}).

\begin{table}[t]
\centering
\resizebox{0.4\columnwidth}{!}{%
\begin{tabular}{@{}lrr@{}}
\toprule
           & Train & Test \\ \midrule
Sarcasm    & 16\%  & 12\% \\
Humblebrag & 38\%  & 49\% \\
Irony      & 15\%  & 11\% \\
Complaints & 14\%  & 10\% \\
Neutral    & 14\%  & 15\% \\
Bragging   & 3\%   & 4\%  \\ \bottomrule
\end{tabular}%
}
\caption{Dataset composition.}
\label{tab:data-composition}
\end{table}

\subsection{Non-Humblebrags}
Humblebrags are often confused with sarcasm and irony, as all three involve an incongruence between the utterance and its intended meaning. To help the model distinguish these phenomena, we included sarcasm and irony as negative samples, alongside direct brags and straightforward complaints. Direct brags convey explicit self-promotion, while complaints reflect surface emotions often present in humblebrags. A more detailed discussion on the differences can be found in \autoref{app: Differences}.

Sarcastic samples were sourced from the SARC dataset \cite{khodak-etal-2018-large}, and ironic ones from SemEval-2018 \cite{van-hee-etal-2018-semeval}. Brags and complaints were taken from \citet{jin-etal-2022-automatic} and \citet{preotiuc-pietro-etal-2019-automatically}, respectively. Neutral sentences, essential for improving class distinction \cite{article}, were drawn from SemEval-2017's sentiment analysis task \cite{rosenthal-etal-2017-semeval}. The dataset composition is shown in \autoref{tab:data-composition}.

\section{Methodology}
\label{Methodology}
We define two task setups: a standard binary classification task, and a sentence completion formulation where humblebrag detection is cast as a Yes/No question answering problem. This enables effective use of decoder-only language models for classification through natural language prompts.
\subsection{Binary Classification for Humblebragging Detection}

In the context of humblebragging detection, the task is to classify a given text \( x \) as either \( C_{\text{HB}} \) (Humblebragging) or \( C_{\text{Non-HB}} \) (Non-humblebragging). The process involves generating text encodings from the input, which are then used for classification. For further details, we refer the reader to \autoref{app:binary}.

\subsection{Classification as a Sentence Completion Task with Yes/No Questions}
Though decoder models are primarily designed for language generation tasks, their ability to predict the next token in a sequence makes them adaptable to various natural language understanding tasks, including classification. Humblebragging classification can be reformulated as a Yes/No question-answering task, where the model determines whether the input text contains humblebragging or not. This approach leverages the natural language understanding capabilities of pre-trained language models to classify text.

\paragraph{Framework}
The input text x is transformed into a prompt structured as

\begin{quote}
\texttt{<definition>}\texttt{<question>}\texttt{<x>}\texttt{<answer>}
\end{quote}

The model is given the \texttt{<question>} along with \texttt{<definition>} and \texttt{<x>} as the input prompt and is expected to generate a text completion for the \texttt{<answer>}. The LLM output in \texttt{<answer>} is then analyzed to determine whether it contains the required word.

\begin{itemize}
\item If \texttt{<answer>} contains \textbf{"Yes"}, the input is classified as y=1 (Humblebragging).
\item If \texttt{<answer>} contains \textbf{"No"}, the input is classified as y = 0 (Non-humblebragging).
\end{itemize}

We evaluate this framework under two settings: \textbf{Z} (zero-shot) and \textbf{Z+D} (zero-shot + definition). In the \textbf{Z} setting, the definition is considered \emph{null}, i.e., no external guidance is provided. In the \textbf{Z+D} setting, the 4-tuple definition is prepended to the input prompt to provide the model with additional context or instruction.

\textbf{Example:}

\begin{quote}
\textbf{Input to LLM (Z+D setting):}

\texttt{<definition>: HB = <B,BT,HM,MT>} \

\texttt{<question>: Is the given text humblebragging or not? Answer in Yes or No only.} \

\texttt{<x>: "Can someone tell the awards committee to chill? Running out of shelf space here!"}

\textbf{Output from LLM:} \

\texttt{<answer>: Yes}

Classification: y=1
\end{quote}

Detailed prompts for both settings are provided in \autoref{app:prompts}.

\section{Experimental Setup}
\label{Experimental Setup}


We conducted experiments with machine learning classifiers, encoder models, decoder models and compared the performance with those of human annotators.

\subsection{Machine Learning Classifiers}  
We evaluated logistic regression and support vector machine (SVM) as simple machine learning based baselines to gauge the task difficulty.

\subsection{Encoder Models}  
Two transformer-based \cite{DBLP:conf/nips/VaswaniSPUJGKP17} encoder models, BERT \cite{devlin-etal-2019-bert} and RoBERTa \cite{liu2019robertarobustlyoptimizedbert}, were evaluated on the task of humblebragging classification. The experiments utilized the \texttt{Adam} optimizer, and 5-fold cross-validation was employed for hyperparameter tuning.

\subsection{Decoder Models}  
For humblebrag detection, decoder models were evaluated in zero-shot (Z) and zero-shot with definition (Z+D) settings, and were also fine-tuned using LoRA (F). In the Z setting, models classify statements as humblebrag or not using only the input. In Z+D, they leverage the 4-tuple definition to guide classification.

Further details about the settings, hyperparameters and human annotators can be found in \autoref{app:exp_setup}.





\section{Results and Discussion}
\label{Results and Discussion}
We present the results of all our experiments in \autoref{tab:results}. The table begins with the majority class baseline, which, in our case, involves predicting every sample as \textit{non-humblebrag}.


\subsection{Quantitative Analysis}

\begin{table*}[ht]
\centering
\resizebox{0.5\textwidth}{!}{%
\begin{tabular}{@{}lrrrr@{}}
\toprule
Model                           & Accuracy & Precision & Recall & F1-Score  \\ \midrule
Baseline                        &0.51   &0.25     &0.50  &0.34     \\
\midrule
Human 1                         & 0.86  & 0.89    & 0.81  & 0.85      \\
Human 2                         & 0.84  & 0.86    & 0.81  & 0.84       \\
Human 3                         & 0.70 & 0.82    & 0.51  & 0.63      \\
Average                          &0.80 &0.86     &0.71   &0.77       \\\midrule
Logistic Regression              & 0.59 & 0.68    & 0.58 & 0.53     \\ 
SVM                              & 0.62 & 0.72    & 0.61 & 0.56      \\
\midrule
BERT-Large-Uncased (F)             & 0.68  & 0.76    & 0.50 & 0.61      \\ 
RoBERTa-Large (F)                   & 0.78 & \textbf{0.91}    & 0.62 & 0.74      \\
\midrule
GPT-4o (Z)                       & 0.84 & 0.78    & 0.94 & 0.85      \\
GPT-4o (Z+D)                     & \textbf{0.89}    & \textbf{0.91}    & 0.85 & \textbf{0.88}      \\
GPT-3.5 (Z)                      & 0.61 & 0.65    & 0.60 & 0.57      \\
GPT-3.5 (Z+D)                    & 0.75 & 0.76    & 0.75 & 0.75      \\
Qwen2.5-7B-Instruct (Z)          & 0.64 & 0.82    & 0.35 & 0.49       \\
Qwen2.5-7B-Instruct (Z+D)        & 0.71 & 0.85    & 0.50 & 0.63      \\
Qwen2.5-7B-Instruct (F)          & 0.67 & 0.85    & 0.40 & 0.54      \\
Mistral-7B-Instruct-v0.3 (Z)      & 0.60    & 0.55    & 0.96 & 0.70      \\
Mistral-7B-Instruct-v0.3 (Z+D)   & 0.60 & 0.55    & 0.96 & 0.70      \\
Llama-3.1-8B-Instruct (Z)        & 0.49 & 0.49    & \textbf{0.99} & 0.66      \\
Llama-3.1-8B-Instruct (Z+D)      & 0.68 & 0.62    & 0.88 & 0.72      \\
Llama-3.1-8B-Instruct (F)        & 0.81 & 0.87    & 0.72 & 0.79      \\
Gemma-1.1-7b-it (Z)              & 0.57 & 0.57    & 0.57 & 0.57      \\
Gemma-1.1-7b-it (Z+D)            & 0.56 & 0.53    & 0.83 & 0.65      \\
Gemma-1.1-7b-it (F)              & 0.71 & 0.71    & 0.44 & 0.60      \\
Vicuna-7b-v1.5 (Z)               & 0.55 & 0.60    & 0.28 & 0.38      \\
Vicuna-7b-v1.5 (Z+D)             & 0.61 & 0.62    & 0.51 & 0.56      \\ \bottomrule
\end{tabular}%
}
\caption{Results of humblebragging classification. Z: zero-shot, Z+D: zero-shot with 4-tuple definition, F: fine-tuned. The best values are in bold.}
\label{tab:results}
\end{table*}

\noindent
Overall, the best-performing model in terms of F1-score is GPT-4o (0.88 F1), surpassing even the best human annotator (0.85 F1). We speculate that this may be due to the extensive linguistic and world knowledge these large-parameter models possess. Among the three human annotators, one performed significantly worse than the others, indicating that the task is non-trivial and can be challenging for some individuals. A detailed discussion on annotation inconsistency is provided in \autoref{app:human-performance}. Moreover, two notable observations emerge from the results.

First, across all decoder models, the Z+D versions consistently outperform their Z counterparts, indicating that our 4-tuple definition effectively aids in detecting humblebragging. To further assess the specific contribution of our definition, we conducted two controlled experiments: (a) replacing the definition in the system prompt with random gibberish (\autoref{app:gibberish}), and (b) substituting our 4-tuple definition with the `textbook definition' (TD) by \citet{wittels2012humblebrag} (\autoref{app:textbook}).

\begin{figure*}[ht]
    \centering
    \includegraphics[width=0.8\linewidth]{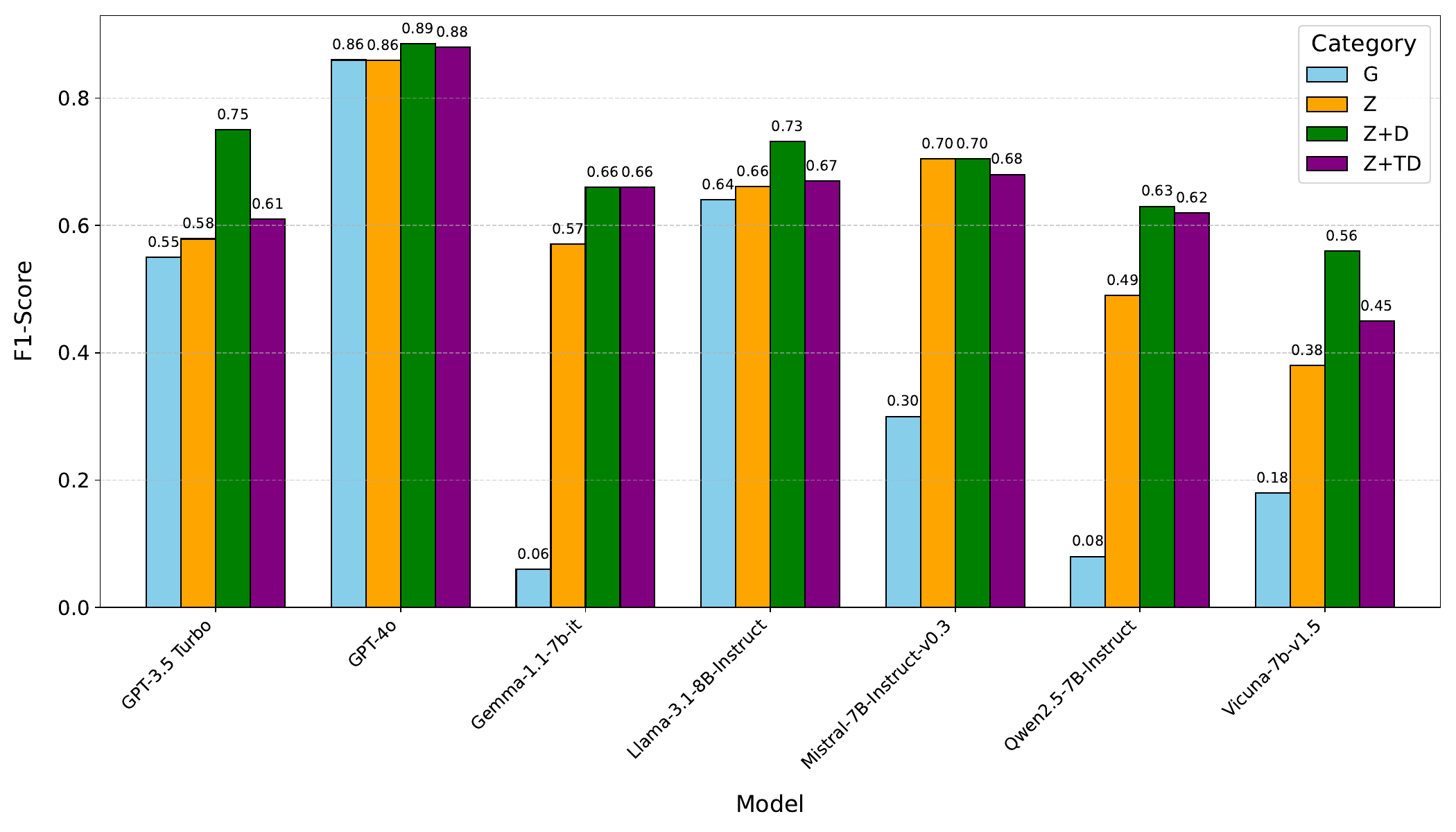}
    \caption{Comparison of F1-scores across scenarios for decoder models: G (Gibberish), Z (Zero-Shot), Z+D (Zero-Shot with our proposed 4-tuple definition of humblebragging), and Z+TD (Zero-Shot with the textbook definition of humblebragging). The Z+D setting achieves the highest F1-scores across all models, demonstrating the effectiveness of the proposed 4-tuple definition in capturing humblebragging nuances.}
    \label{fig:barchart}
\end{figure*}

\noindent
The results, summarized in \autoref{fig:barchart}, clearly demonstrate that models prompted with our 4-tuple definition perform better than both the gibberish and textbook alternatives, reaffirming its utility in enhancing humblebrag classification.

Second, fine-tuning with our HB-24 dataset improved the F1-scores of the majority of the models. Both encoder models and three decoder models (Llama, Gemma, and Qwen) showed significant gains from fine-tuning. Interestingly, fine-tuned RoBERTa outperformed all 7–8 billion parameter decoder models except for Llama (F). This highlights the superior classification capabilities of encoder-only models when a high-quality dataset for fine-tuning is available.  

\begin{figure*}[ht]
     \centering
     \begin{subfigure}[b]{0.3\linewidth}
         \centering
         \includegraphics[width=\linewidth]{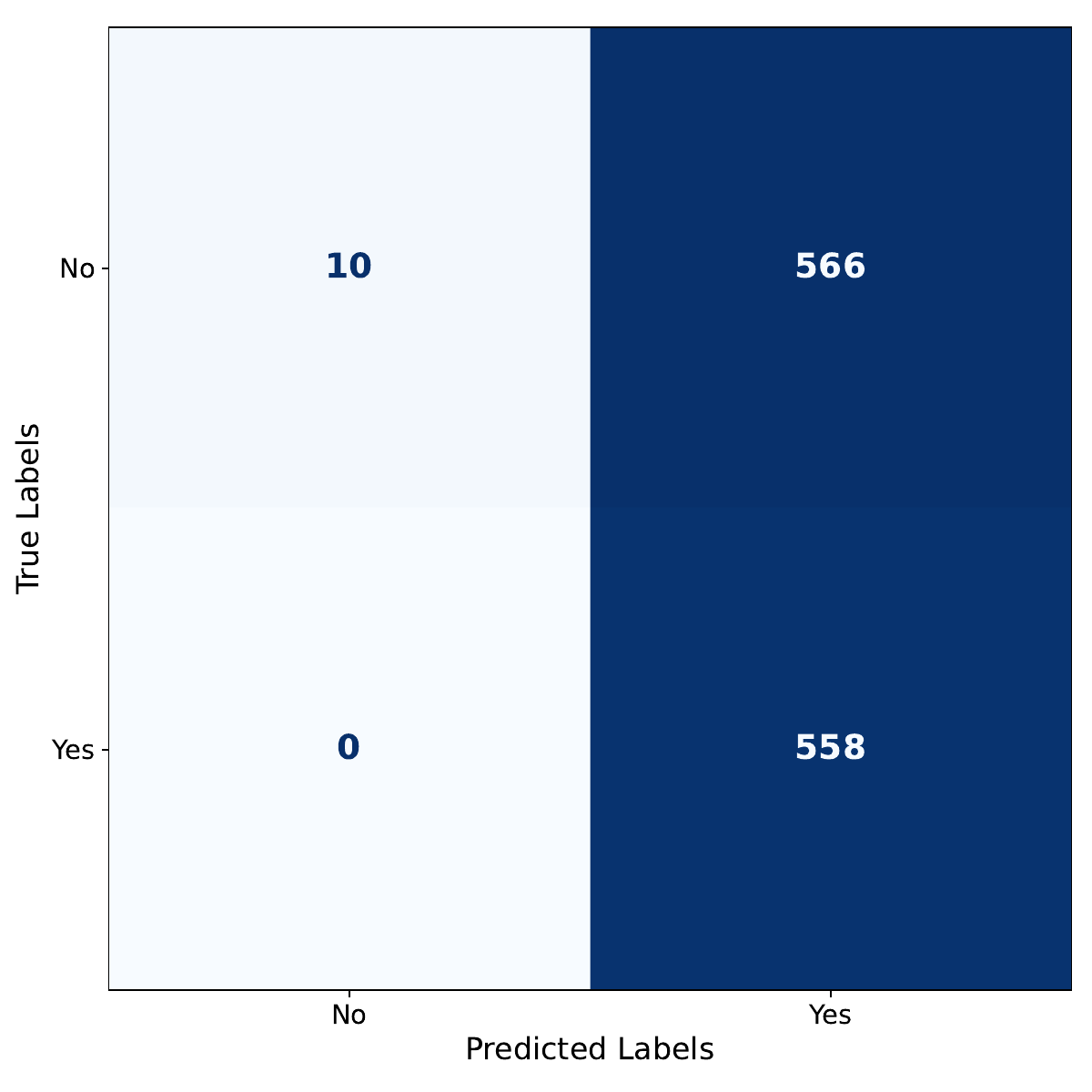}
         \caption{Llama (Z)}
         \label{fig:llama_Z_main}
     \end{subfigure}
     \hfill
     \begin{subfigure}[b]{0.3\linewidth}
         \centering
         \includegraphics[width=\linewidth]{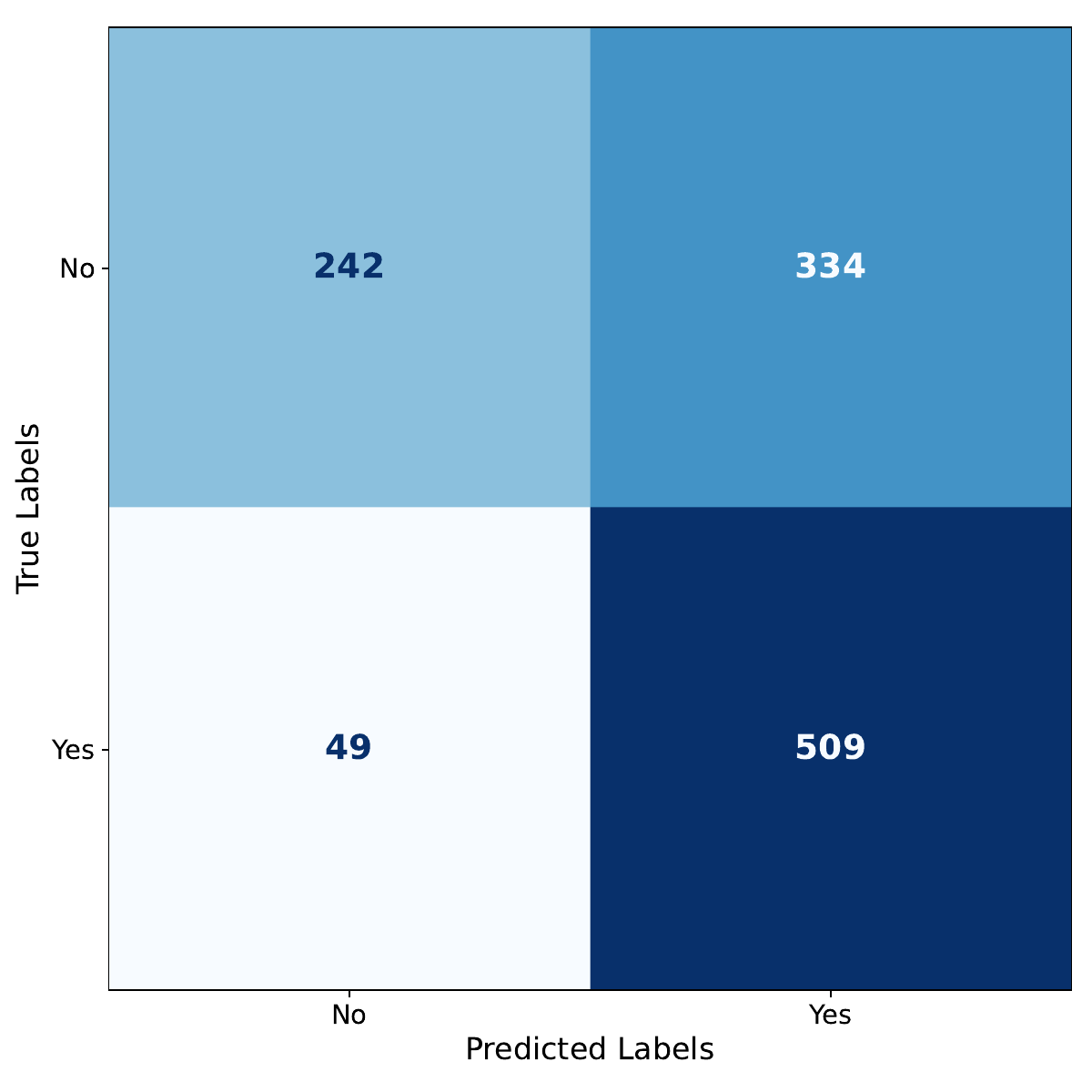}
         \caption{Llama (Z+D)}
         \label{fig:llama_ZD_main}
     \end{subfigure}
     \hfill
     \begin{subfigure}[b]{0.3\linewidth}
         \centering
         \includegraphics[width=\linewidth]{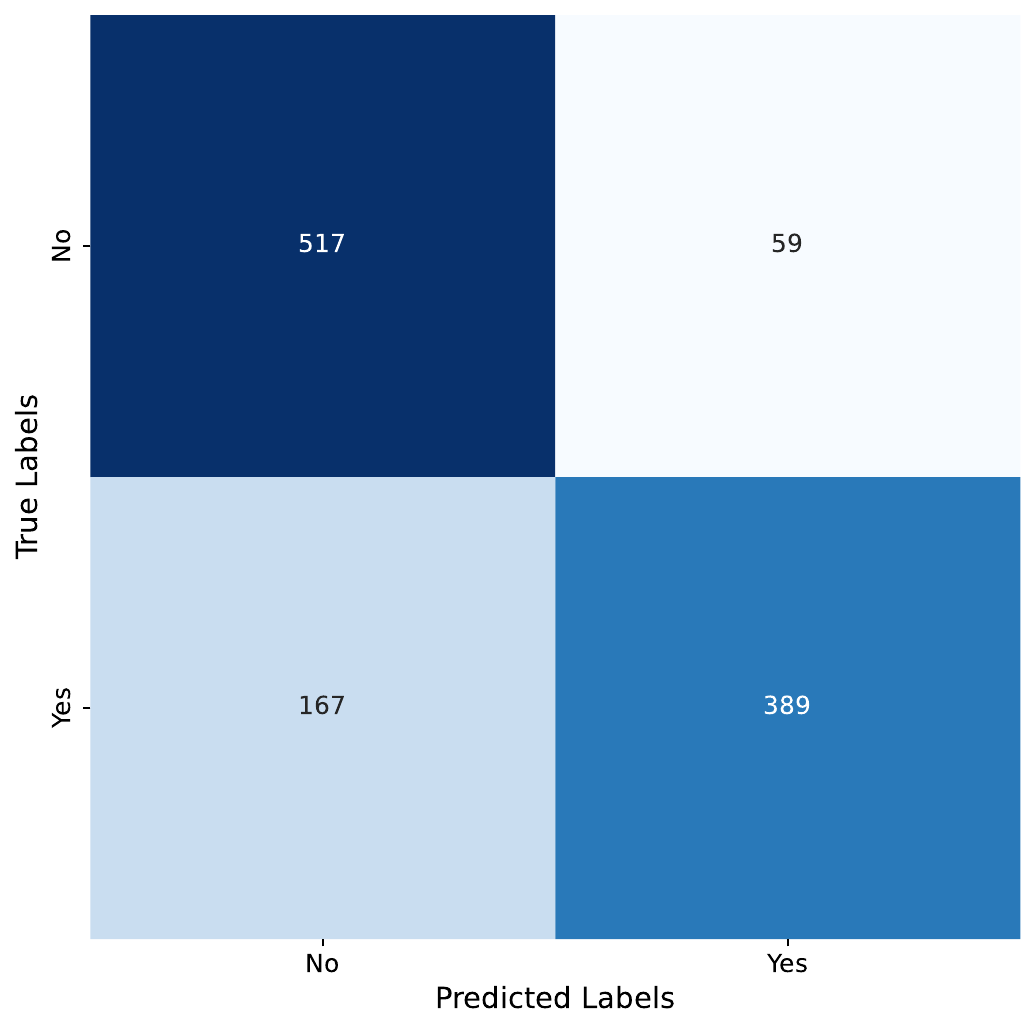}
         \caption{Llama (F)}
         \label{fig:llama_F_main}
     \end{subfigure}
        \caption{Confusion matrices for Llama Z vs Z+D vs F. Gradual improvement in the model can be observed from left to right.}
        \label{fig:confusion-main}
\end{figure*}

We illustrate in \autoref{fig:confusion-main} Llama's progression from an F1-score of 0.66 in the zero-shot setting (Z) to an F1-score of 0.79 after fine-tuning (F), through confusion matrices. Llama (Z) primarily predicted the \textit{yes} label for almost all samples. With our 4-tuple definition in Llama (Z+D), the model began to identify non-humblebrags, bringing more balance to the confusion matrix. After fine-tuning, Llama (F) became more proficient in identifying non-humblebrags while sacrificing some true positives. Confusion matrices of other models can be found in \autoref{app:confusion}.

Lastly, we observed anomalous behavior with Mistral and Vicuna, where fine-tuning led to lower F1-scores. Notably, Mistral's zero-shot performance already exceeded that of other models in its category, including the larger GPT-3.5 and any further fine tuning is resulting in catastrophic forgetting. In Vicuna's case, the fine-tuned model produced random texts and emojis and was extremely sensitive to slight prompt changes in the Z+D setting, requiring removal of the final sentence from the system prompt (\autoref{app:prompts}). Additional insights on performance degradation are provided in \autoref{app:degradation}.

\subsection{Qualitative Analysis}

In this section, we analyze cases of agreement and disagreement between human annotators and models.

All humans and models correctly classified the following as humblebrags, as the brags were easy to identify with a clear distinction between the brag and the mask segments:

\begin{quote}
    T1: \textit{In the limo riding to airport. Sucks being alone though}

    T2: \textit{just tried to pre-order my book. couldnt figure it out. did anyone try?}
\end{quote}

\noindent
For the following humblebrags, humans classified them as \textit{no}, while models classified them as \textit{yes}:

\begin{quote}
    T3: \textit{I forget. What airport do u fly into to get to Maui?}

    T4: \textit{I just had my first screaming girl encounter. She probably had me confused withsomeone else.}
\end{quote}

\noindent
In the case of T3, the disagreement could stem from a lack of knowledge regarding Maui as an exotic travel destination. On the other hand, the phrase ``screaming girl'' in T4 might not have been understood by the annotators due to cultural differences. Some cultures might interpret the phrase at its surface level without delving into its deeper meaning.

\noindent
For the following humblebrag, humans said \textit{yes}, while models said \textit{no}:

\begin{quote}
    T5: \textit{The CNN-LA green room is a cold and lonely place at 7 on a Sunday morning.Funnily enough, CNN LA green room a cold and lonely place at 10 on a Monday too.}
\end{quote}

\noindent
We hypothesize that the model might have been confused by the incongruity between ``cold and lonely'' and ``Funnily enough,'' interpreting it as sarcasm instead of a humblebrag.

\noindent
For the following non-humblebrag, both humans and models classified it as \textit{yes}:

\begin{quote}
    T6: \textit{i decided to become my own boss to have more free time.. now i have no time left whatsoever.}
\end{quote}

\noindent
T6 is a rare case where our assumption that humblebrags are not present in the datasets used to create our negative samples was violated. After encountering this example, we reviewed our dataset again to ensure no other such case exists.

\noindent
For the following non-humblebrag, models said \textit{no}, but humans said \textit{yes}:

\begin{quote}
    T7: \textit{After years of fumbling around , I have finally found a skin care product that works for me . Well , at least for now ?}

    T8: \textit{TheoCorleone david\_maclellan Shit! I better shut my stupid girly mouth because im so concerned about what men might think of me.}
\end{quote}

\noindent
T7 and T8 represent classic cases of annotator bias, where annotators attempt to imagine a non-existent context and incorrectly classify the samples as belonging to the positive class. This bias arises because annotators, subconsciously influenced by their task, oversearch for humblebrags in the data as they are tasked to annotate for humblebragging detection.

While our qualitative analysis highlights the nuanced and often subjective nature of humblebrag detection, it also reveals patterns in how models and humans interpret brag-masking cues utilizing our 4-tuple framework. In the next subsection, we compare our framework with another plausible alternative to further reaffirm its suitability for the task.

\subsection{4-tuple vs Sentiment Opposition Model}
We pit our 4-tuple framework against the Sentiment Opposition Model (SOM) (\autoref{app:som}), which is based on detecting incongruity between surface sentiment and intended sentiment. For instance when a statement appears negative or modest but actually conveys an underlying brag.

\autoref{tab:llama_zd_som_comparison} indicates that the 4-tuple Definition (Z+D) outperforms the Sentiment Opposition Model (Z+SOM) in detecting humblebragging, achieving a higher recall and F1-score while maintaining comparable precision. This suggests that the 4-tuple framework more effectively captures both the brag and its masking component, leading to better overall detection. While SOM improves precision by reducing false positives, it sacrifices recall, making it less sensitive to subtle humblebraggings. The higher F1-score of the 4-tuple model confirms its better overall balance between precision and recall, making it a more robust approach for humblebragging detection. 

\begin{table}[t]
\centering
\resizebox{0.5\linewidth}{!}{%
\renewcommand{\arraystretch}{1.2} 
\setlength{\tabcolsep}{6pt} 
\begin{tabular}{lcccc} 
\toprule
\textbf{Model} & \textbf{A} & \textbf{P} & \textbf{R} & \textbf{F1} \\ 
\midrule
Z+D & \textbf{0.68} & 0.62 & \textbf{0.88} & \textbf{0.72} \\ 
Z+SOM & 0.66 & \textbf{0.64} & 0.74 & 0.68 \\ 
\bottomrule
\end{tabular}
}
\caption{Comparison of Llama-3.1-8B-Instruct using the 4-tuple Definition (Z+D) vs. the Sentiment Opposition Model (SOM). A: Accuracy, P: Precision, R: Recall, F1: F1-score.}
\label{tab:llama_zd_som_comparison}
\end{table}

\subsection{Humblebragging Component Identification}

We go a step further in testing the ability of models in identifying the brag and mask components in a given humblebrag by applying our definition. This structured interpretation of humblebragging text demonstrates the practical usefulness of our definition in capturing the nuances of humblebragging enabling models to distinguish the underlying brag from its masked presentation, making detection more accurate and interpretable. \autoref{app:interpretation} showcases how different language models process humblebragging using this framework, highlighting variations in their ability to correctly segment and classify such statements. This analysis reinforces the effectiveness of our definition in improving both automated detection and human understanding of humblebragging in natural language.

\subsection{Impact of Humblebragging Detection on Downstream Applications}

To evaluate the utility of humblebragging detection in a downstream task, we conducted an intended-polarity classification experiment. From the HB-24 test set, we filtered out irony and neutral cases and defined gold labels as follows: humblebragging and bragging were labeled positive, while sarcasm and complaints were labeled negative.

As a baseline, we used RoBERTa-large fine-tuned on the SST-2 dataset \cite{socher-etal-2013-recursive}, referred to as R-SST2. We then introduced a GPT-based classifier to detect humblebragging and sarcasm, adjusting the sentiment scores accordingly: +1 for humblebrags (to reflect their underlying positivity) and –1 for sarcasm (to correct for overstated positivity). The modified classification module, which integrates these adjustments into the R-SST predictions, is referred to as R-HBSC (HumbleBragging and SarCasm).
\begin{table}[t]
    \centering
    \resizebox{0.6\linewidth}{!}{%
    \renewcommand{\arraystretch}{1.2} 
    \setlength{\tabcolsep}{8pt} 
    \begin{tabular}{lcccc} 
        \toprule
        \textbf{Model} & \textbf{A} & \textbf{P} & \textbf{R} & \textbf{F1} \\ 
        \midrule
        R-SST2 & 0.53 & 0.69 & 0.53 & 0.51 \\ 
        R-HBSC & \textbf{0.82} & \textbf{0.86} & \textbf{0.82} & \textbf{0.83} \\ 
        \bottomrule
    \end{tabular}
    }
    \caption{Comparison of vanilla RoBERTa-SST2 (R-SST2) with RoBerta-HBSC (R-HBSC). A: Accuracy, P: Precision, R: Recall, F1: F1-score.}
    \label{tab:roberta_hb_sc}
\end{table}

\autoref{tab:roberta_hb_sc} shows that adding this pragmatic layer significantly improves sentiment classification across all metrics. This demonstrates that accounting for implicit cues like humblebragging and sarcasm better aligns model predictions with intended sentiment. 


\section{Conclusion and Future Work}
\label{Conclusion and Future Work}

We introduced the task of automatic humblebragging detection, formalized through our proposed 4-tuple definition. We benchmarked various machine learning, deep learning, and large language models on this task, providing a comparative analysis against human performance. We also demonstrate that our 4-tuple definition significantly improves the zero-shot capabilities of all decoder models. Additionally, we released a synthetic dataset, HB-24, generated using GPT-4o, to facilitate further research. Our experiments and analysis reveal that detecting humblebragging is a challenging task, even for humans. This study lays the groundwork for exploring this intricate linguistic phenomenon and its integration into natural language understanding systems.

Future research could aim to enhance models for identifying humblebragging, fostering a deeper comprehension of this distinct communication style. This may include methods for utilizing contextual cues more effectively. Another valuable direction could involve generating humblebrag captions for images. Additionally, machines could be trained to transform direct brags into humblebrags.

\section*{Limitations}
The inherent subjectivity of humblebragging complicates the creation of universally agreed-upon labels, as even humans often struggle to classify such statements consistently. Additionally, while machine-generated texts are sophisticated and well-structured, they often lack the spontaneity and imperfections typical of human-authored texts. For instance, the model's inability to use certain casual or curse words, as well as elongated words like \textit{sooooo} or \textit{goood}, which are often present in human-written humblebrags. This creates a mismatch when using synthetic datasets like HB-24, which, despite being a valuable resource, may fail to fully capture the linguistic diversity and subtleties of real-world humblebragging, thereby limiting the generalizability of trained models. Moreover, the task itself remains underexplored, with no prior benchmarks or resources, making it difficult to contextualize results within the larger field of natural language understanding.


\bibliography{anthology,custom}

\appendix

\section{On the Prevalence of Humblebragging}
\label{app: Prevalence}
\begin{figure*}
    \centering
    \includegraphics[width=0.5\linewidth]{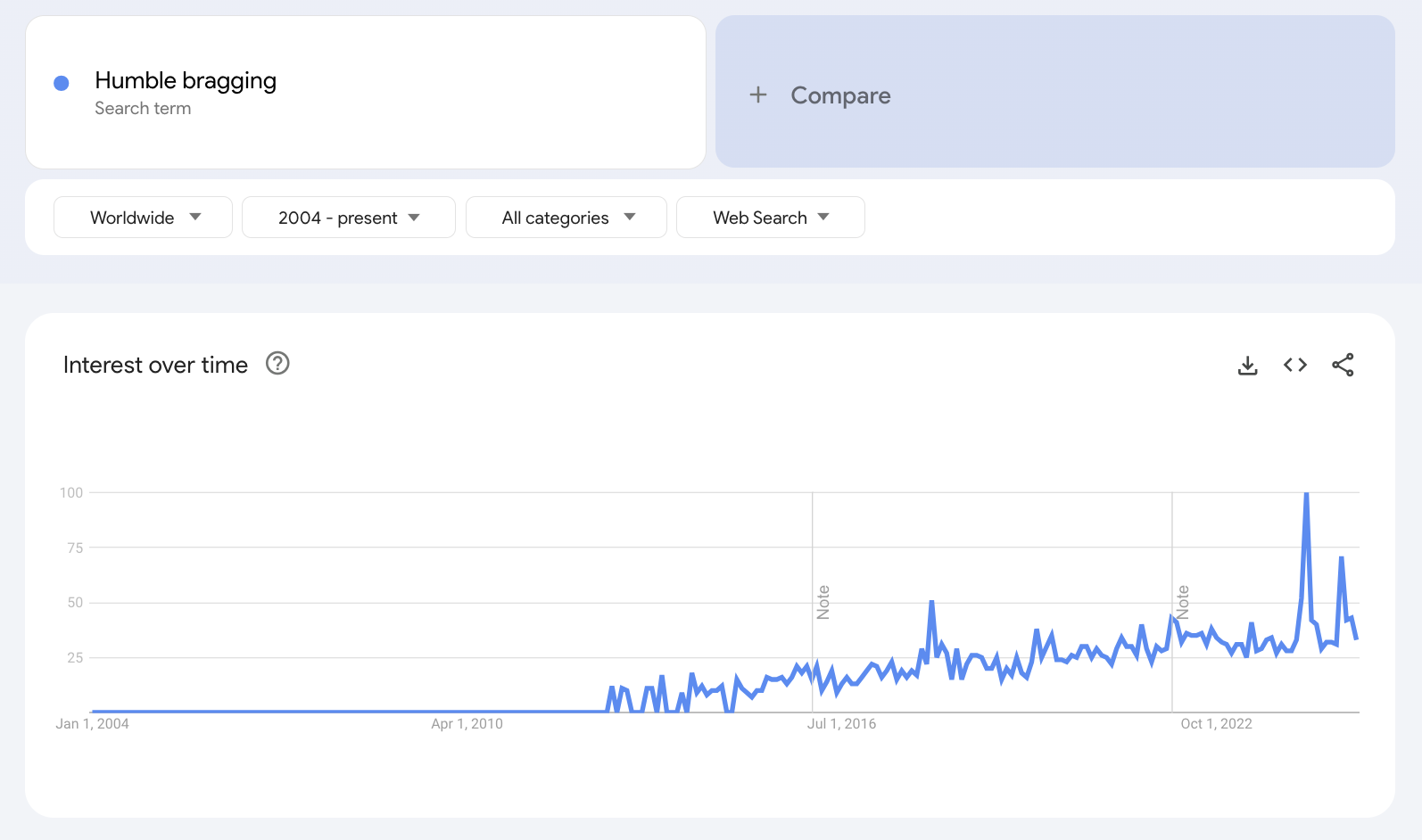}
    \caption{Polpularity of humblebragging.}
    \label{fig:trends}
\end{figure*}

Although less common compared to sarcasm or irony, humblebragging is not nonexistent, as evidenced by the Google Trends graph reproduced in \autoref{fig:trends}.

The graph clearly shows that the popularity of the term has been increasing gradually over time.

Moreover Reddit has a dedicated r/humblebrag subreddit with 188 thousand members. In contrast r/sarcasm has only 39 thousand members while r/irony has only 60 thousand.

In case of Twitter/X, tweetbinder.com returns the following tweet counts over the past week (search performed on 14 Feb 2025):
\begin{itemize}
    \item \#humblebrag: 72
    \item \#sarcasm: 200
    \item \#irony: 200
\end{itemize}
Note here that 200 is the limit for the free tier searches.

All of the above statistics correspond to the explicit mentions of the phenomena in question. We suspect implicit humblebragging is much more common in social media. But providing statistics about it is all the more difficult.

Additionally, as observed by Wittels in his book \cite{wittels2012humblebrag}, the phenomena of humblebragging is not recent. In fact only the term for the phenomena is recent. People have been engaging in humblebragging from historical times knowingly or unknowingly.

\section{How Humblebragging Differs from Irony, Sarcasm, Bragging and Complaint}
\label{app: Differences}

While irony, sarcasm, and humblebragging rely on indirect communication, bragging and complaints are more direct. Irony presents a contrast between expectation and reality, often meaning the opposite of what is stated. Sarcasm builds on irony but with a sharper, mocking tone, where the surface meaning appears positive, but the intent is negative. Humblebragging disguises self-promotion within a complaint or self-deprecating remark, appearing negative while aiming to impress.

In contrast, bragging and complaints do not rely on hidden meanings. Bragging openly expresses pride, maintaining a positive tone in both surface meaning and intent. Complaints directly convey dissatisfaction, with both their surface and intended polarity being negative. These distinctions are summarized in \autoref{tab:polarity}.

\begin{table*}[ht]
    \centering
    \begin{tabular}{@{}lll@{}}
        \toprule
        \textbf{Phenomenon} & \textbf{Surface Polarity} & \textbf{Intended Polarity} \\ \midrule
        Irony               & Positive or Negative      & Opposite of surface polarity \\
        Sarcasm             & Positive                 & Negative                     \\
        Humblebragging      & Negative                 & Positive                     \\
        Bragging            & Positive                 & Positive                     \\
        Complaint           & Negative                 & Negative                     \\ \bottomrule
    \end{tabular}
    \caption{Surface and Intended Polarity of Different Phenomena}
    \label{tab:polarity}
\end{table*}

\section{Humblebrag Categories}
\label{apx:categories}

The mapping between Wittel’s humblebrag themes to Sezer's humblebrag categories is shown in \autoref{table:categorized_humblebragging}.

\begin{table*}[ht] 
\centering
\begin{tabular}{@{}p{7cm}p{7cm}@{}}
\toprule
\textbf{Category} \cite{2017-40996-001} & \textbf{Theme} \cite{wittels2012humblebrag} \\ \midrule
Looks and Attractiveness & 
Ugh, Being Hot Sure Can Be Annoying! \newline
Ugh, It’s Tough Being a Model \newline
Ugh, I’m Too Skinny! \newline
Ugh, People Keep Hitting on Me! \\
\midrule
Achievements & 
Ugh, Can You Believe They Included Me on This List? \newline
Ugh, I Can’t Believe I Won an Award\\
\midrule
Performance at Work & 
Ugh, I’m So Successful \\
\midrule
Money and Wealth & 
Ugh, I Hate Having All This Money! \\
\midrule
Intelligence & 
Ugh, I’m a Genius \\
\midrule
Personality & 
Ugh, I’m So Humble! \newline
Ugh, It’s Hard Being So Charitable! \\
\midrule
Social Life & 
Ugh, It’s So Weird Getting Recognized! \newline
Ugh, I Hate People Wanting My Picture and Autograph All the Time \newline
Ugh, I’m at an Exclusive Event! \newline
Ugh, Being an Author Is Hard! \newline
Ugh, How’d I Get Here??? How Is This My Life??? \newline
Ugh, I Travel Too Much! \\
\midrule
Miscellaneous & 
Ugh, I Can’t Believe I Was Mentioned in This Thing! \\

\bottomrule
\end{tabular}
\caption{Mapping Wittel's humblebrag themes to Sezer et al.'s humblebrag categories.}
\label{table:categorized_humblebragging}
\end{table*}

\section{Data Generation Prompts}
\label{app:data-prompts}

\subsection*{General Prompt}
\begin{lstlisting}
You are now a person about to humblebrag 
about your recent achievement to attract 
people's attention and make them praise 
you.But you can't state the obvious. You 
have to present it in such a way that it 
sounds like a complaint without reducing 
the importance of the achievement. 
There should be a strong incongruence. 
Make sure these are tweets, and keep the
tone casual. Be specific about your 
achievements and use diverse topics. 
Do not use topics already generated, 
and do not follow a pattern for 
beginning the text.
\end{lstlisting}

\subsection*{Prompt with Themes}
\begin{lstlisting}
Here is the definition of humblebragging: 
a specific type of brag that masks the 
boasting part of a statement in a 
faux-humble guise. The false humility 
allows the offender to boast about their 
"achievements" without any sense of shame 
or guilt. Humblebrags are usually 
self-deprecating in nature. 

Now, you are a person who is about to 
humblebrag on Twitter with the theme 
<theme> and it should sound casual. 
Use the above definition and generate 
humblebrags.
\end{lstlisting}

\section{Dataset Quality Assurance}
\label{app:Dataset Quality Assurance}

After synthetic data generation, a manual verification was performed by the first two authors of the paper. In this manual verification step, the main aim was to ensure the selection of high-quality samples for the dataset by:

\begin{itemize}
    \item Removal of near duplicates.
    \item Inclusion of diverse categories of humblebrags that represent real-world scenarios.
\end{itemize}

Moreover, the 4-tuple definition of humblebragging guided the entire manual filtering stage. For instance, out of the following samples, only one was selected, and the rest were discarded:

\begin{itemize}
    \item "Why can’t they just serve normal snacks in first class? Caviar and champagne get so repetitive."
    \item "Why do they always offer turn-down service on long-haul flights? Sometimes I just want to make my own bed."
    \item "Why does the in-flight chef always insist on preparing gourmet meals? Sometimes I just crave a simple sandwich."
    \item "Why do first-class cabins have private suites? I kind of miss the open seating vibe of economy."
\end{itemize}

This manual filtering stage was followed by a discussion round wherein the two authors discussed both the filtered-in and filtered-out samples. For cases of disagreement, another round of filtering followed by discussion was performed.

\section{Binary Classification Framework for Humblebragging Detection}
\label{app:binary}
\paragraph{Text Encoding Generation} The input text \( x \) is converted into a numerical representation \( \mathbf{e} \) using any of the available encoding techniques. Formally:
\begin{equation}
\mathbf{e} = f_{\text{encoder}}(x)
\end{equation}

where:
\begin{itemize}
    \item \( x \): Input text.
    \item \( \mathbf{e} \): Encoded representation of the text, typically a fixed-dimensional vector.
    \item \( f_{\text{encoder}} \): The encoding function, such as TF-IDF, BERT or a similar pre-trained transformer.
\end{itemize}

This encoded representation \( \mathbf{e} \) captures semantic and contextual information from the input text, enabling effective classification.

\paragraph{Binary Classification} Using the encoded representation \( \mathbf{e} \), the model predicts the probability \( \hat{y} \in [0, 1] \) for the text belonging to the class \( C_{\text{HB}} \). The true label \( y \) is \( y = 1 \) for humblebragging and \( y = 0 \) otherwise. 

The Binary Cross-Entropy (BCE) loss for this task is defined as:

\begin{equation}
\resizebox{0.9\hsize}{!}{$
\mathcal{L}_{\text{BCE}}(y, \hat{y}) = - \frac{1}{N} \sum_{i=1}^{N} \Big( y_i \log(\hat{y}_i) + (1 - y_i) \log(1 - \hat{y}_i) \Big)
$}
\end{equation}

where:
\begin{itemize}
    \item \( N \): Total number of samples in the dataset.
    \item \( y_i \): True label for the \( i \)-th sample (\( y_i \in \{0, 1\} \)), where \( y_i = 1 \) indicates a humblebrag.
    \item \( \hat{y}_i \): Predicted probability that the \( i \)-th sample is a humblebrag (\( \hat{y}_i \in [0, 1] \)).
\end{itemize}

\paragraph{Objective} The model is trained to minimize \( \mathcal{L}_{\text{BCE}} \) over the dataset, improving its ability to accurately classify texts as humblebragging or non-humblebragging.

\section{Inference Prompts}
\label{app:prompts}

\subsection*{User Prompt}
\begin{lstlisting}
### Question: Is this a humble brag? 
Answer in yes or no only.
### Statement: {data_point['text']}
### Answer:
\end{lstlisting}

\subsection*{System Prompt}
\label{app:system-prompt}
\begin{lstlisting}
A humble brag comprises the following 
components:

1. Brag:
   - The segment of the text that 
   explicitly conveys the act of 
   bragging.

2. Brag Theme:
   - The overarching theme or specific
   category of the brag embedded within 
   the statement.
   - Possible categories include:
       - Looks and Attractiveness 
       - Achievements 
       - Performance at Work 
       - Money and Wealth 
       - Intelligence 
       - Personality 
       - Social Life 
       - Miscellaneous

3. Humble Mask:
   - The element of the text that adopts 
   a modest or self-deprecating tone to 
   obscure or mitigate the act of 
   bragging.

4. Mask Type:
   - Specifies whether the humble mask 
   adopts a modest tone or a 
   complaining approach.

Now you are about to classify if a given 
sentence is a humble brag or not using 
the above definition.
\end{lstlisting}

\section{Detailed Experimental Setup}
\label{app:exp_setup}

Below we outline the hyperparameter settings used for various models along with a brief discussion of the human annotation.

\subsection{Machine Learning Classifiers}  

We conducted a grid search on various hyperparameters to identify the best combination for each classifier. For the Logistic Regression model, we set the maximum number of iterations (\texttt{max\_iter}) to 100 to ensure convergence, the regularization strength (\texttt{C}) to 0.1 to control overfitting, and a fixed random state (\texttt{random\_state=42}) for reproducibility. For the Support Vector Classifier (SVC), we used a radial basis function (RBF) kernel (\texttt{kernel=`rbf'}) to capture non-linear relationships in the data and set the regularization parameter (\texttt{C}) to 10.

Unlike other types of textual data, tweets are often informal and unique in their composition. They frequently include \textit{emojis} and \textit{emoticons}, which add emotional or contextual cues. Additionally, tweets commonly feature elongated words (e.g., \textit{soooo} or \textit{goood}) and repeated characters for emphasis or emotional expression. While modern tokenizers utilized by pre-trained networks are designed to handle these emojis and elongated words effectively, traditional machine learning algorithms often struggle with such unconventional text patterns.

Thus, for machine learning classifiers, we incorporated existing pre-processing techniques in two distinct phases. In the first phase, we replaced all emojis with their corresponding verbal explanations, as suggested by \citet{singh-etal-2019-incorporating}, to retain the semantic information. In the second phase, we utilized \textit{ekphrasis} \cite{baziotis-pelekis-doulkeridis:2017:SemEval2}, a specialized text pre-processing tool, to handle hashtags, elongated and repeated words, URLs, and numeric information. The output was further pre-processed by removing stop words and punctuation, constructing unigram and bigram tokens, limiting the vocabulary to the top 10,000 tokens by frequency, and requiring each token to appear in at least two training samples. The resulting numerical representations, created using TF-IDF, were used as inputs for machine learning classifiers- Support Vector Machine and Logistic Regression.

\subsection{Encoder Models}   

BERT-Large-Uncased (340M) was trained with a learning rate of 5e-3, batch size of 16, and 4 epochs, while RoBERTa-Large (355M) used a learning rate of 5e-4, batch size of 32, and 5 epochs. Both models shared a maximum sequence length of 128, a weight decay of 0.01, a warmup ratio of 0.1, and gradient clipping at 1.0.

\subsection{Decoder Models}  

We evaluated open-source models from Hugging Face—Qwen2.5-7B-Instruct \cite{qwen2}, Llama 3.1-8B-Instruct \cite{grattafiori2024llama3herdmodels}, Gemma 1.1 7B (IT) \cite{gemmateam2024gemmaopenmodelsbased}, Mistral-7B-Instruct-v0.3 \cite{jiang2023mistral7b}, and Vicuna-7B-v1.5 \cite{zheng2023judging}—as well as GPT-3.5 and GPT-4o via OpenAI’s API\footnote{\url{https://platform.openai.com/docs/overview}}. Outputs were limited to two tokens, and each model was run five times to record average metrics. Prompt details are in \autoref{app:prompts}.

Fine-tuning was conducted using LoRA (Low-Rank Adaptation) with a scaling factor (\texttt{lora\_alpha}) of 8, rank (\texttt{r}) of 16, and targeted attention modules (\texttt{q\_proj} and \texttt{k\_proj}). The dataset was split into 80\% training and 20\% validation with a random seed of 42. Training employed a cosine learning rate scheduler with a learning rate of 1e-5, weight decay of 0.01, and 10 warmup steps, over 4 epochs with a batch size of 8 and gradient accumulation steps of 4. The maximum sequence length was set to 512 tokens, and the \texttt{SFTTrainer} was utilized for efficient fine-tuning.  

All experiments were conducted on NVIDIA A100-SXM4-80GB GPUs, utilizing approximately 300 GPU hours in total.

\subsection{Human Performance}
To evaluate human performance, the test dataset was labeled by three independent annotators, including two with Masters degrees in Linguistics and Arts, and a final-year Masters student in Computer Applications. All were proficient in English and experienced in professional annotation. An initial meeting with the authors covered annotation guidelines and the scope of the task. The annotators first labeled a small subset of the dataset, and disagreements were discussed to ensure consistency. They then labeled the entire test set as \textit{yes} or \textit{no}, indicating the presence or absence of humblebragging. The annotators were compensated according to University norms.

\section{Inconsistency in Human Annotation}
\label{app:human-performance}
On analyzing the misclassifications made by the lowest-performing human annotator, we observed that the primary factor was not the 4-tuple definition itself. Instead, cultural differences between the source of the tweets and the annotator’s background played a significant role, leading to missed contextual cues.

For instance, the following examples were misclassified by the annotator as they failed to identify celebrity status of people mentiond in the example:
\begin{itemize}
    \item \textit{Sitting next to Penny Marshall at the Lakers Game. \#GEEKINGOUT}
    \item \textit{Tonight: private dinner/event by Miles Davis Estate---his 85th birthday with his family, musicians, media (Beyond humbled/honored to be invited).}
    \item \textit{Watching myself on Larry King. Achievement diminished in ad break by catheter commercial.}
\end{itemize}

\noindent
To address this issue, future studies should consider recruiting annotators from diverse cultural and geographical backgrounds or implementing mechanisms to provide additional contextual information. However, such interventions are beyond the scope of the present study and are proposed as directions for future research.

\section{Random Gibberish Prompt}
\label{app:gibberish}

\begin{lstlisting}
A Quantum Pancake involves the following 
components:
    
    1. Flapjack Fluctuation: 
       - Analyze the positive or negative 
       curvature of the syrupy timeline 
       as the pancake flips through 
       space-time.
       
    2. Stack Dynamics: 
       - Identify the structural 
       integrity of the pancake layers 
       and their inter-dimensional 
       fluffiness coefficient.
    
    3. Butter Singularities: 
       - Highlight the concentrated 
       points of creamy chaos where the 
       butter both exists and does not 
       exist simultaneously.
    
    4. Maple Entanglement: 
       - Describe the sticky phenomena 
       where the syrup defies Newtonian 
       logic to connect pancakes across 
       parallel brunch universes.

Now you are about to classify if a given 
sentence is a humble brag or not using 
the above definition.
\end{lstlisting}

\section{Text Book Definition}
\label{app:textbook}

\begin{lstlisting}
A specific type of brag that masks 
the boasting part of a statement 
in a faux-humble guise. The false 
humility allows the offender to 
boast about their achievements
without any sense of shame or guilt. 
Humblebrags are usually 
self-deprecating in nature.

\end{lstlisting}

\section{Confusion Matrices}
\label{app:confusion}

\begin{figure*}[t]
     \centering
     \begin{subfigure}[b]{0.3\linewidth}
         \centering
         \includegraphics[width=\linewidth]{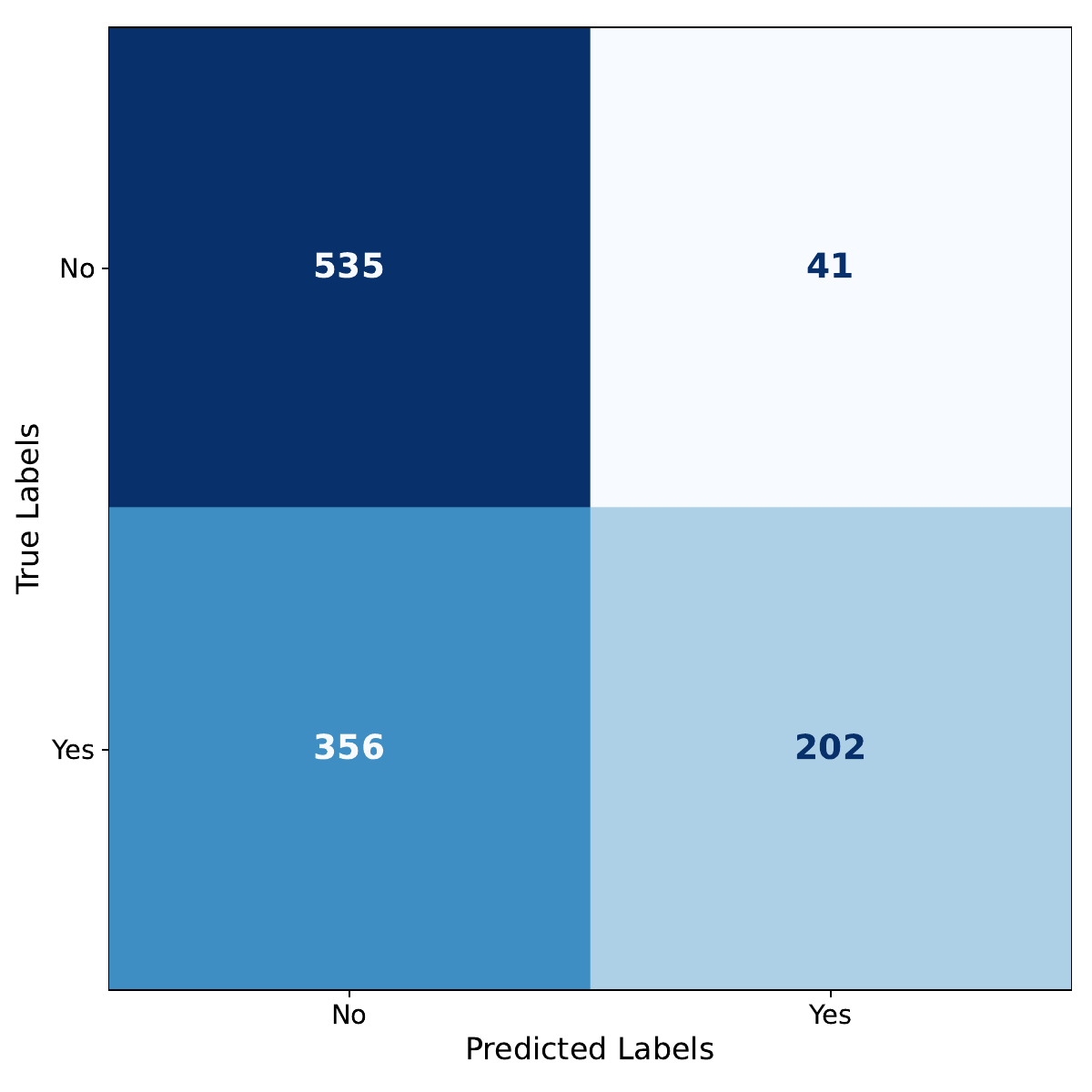}
         \caption{Qwen (Z)}
         \label{fig:qwen_Z}
     \end{subfigure}
     \hfill
     \begin{subfigure}[b]{0.3\linewidth}
         \centering
         \includegraphics[width=\linewidth]{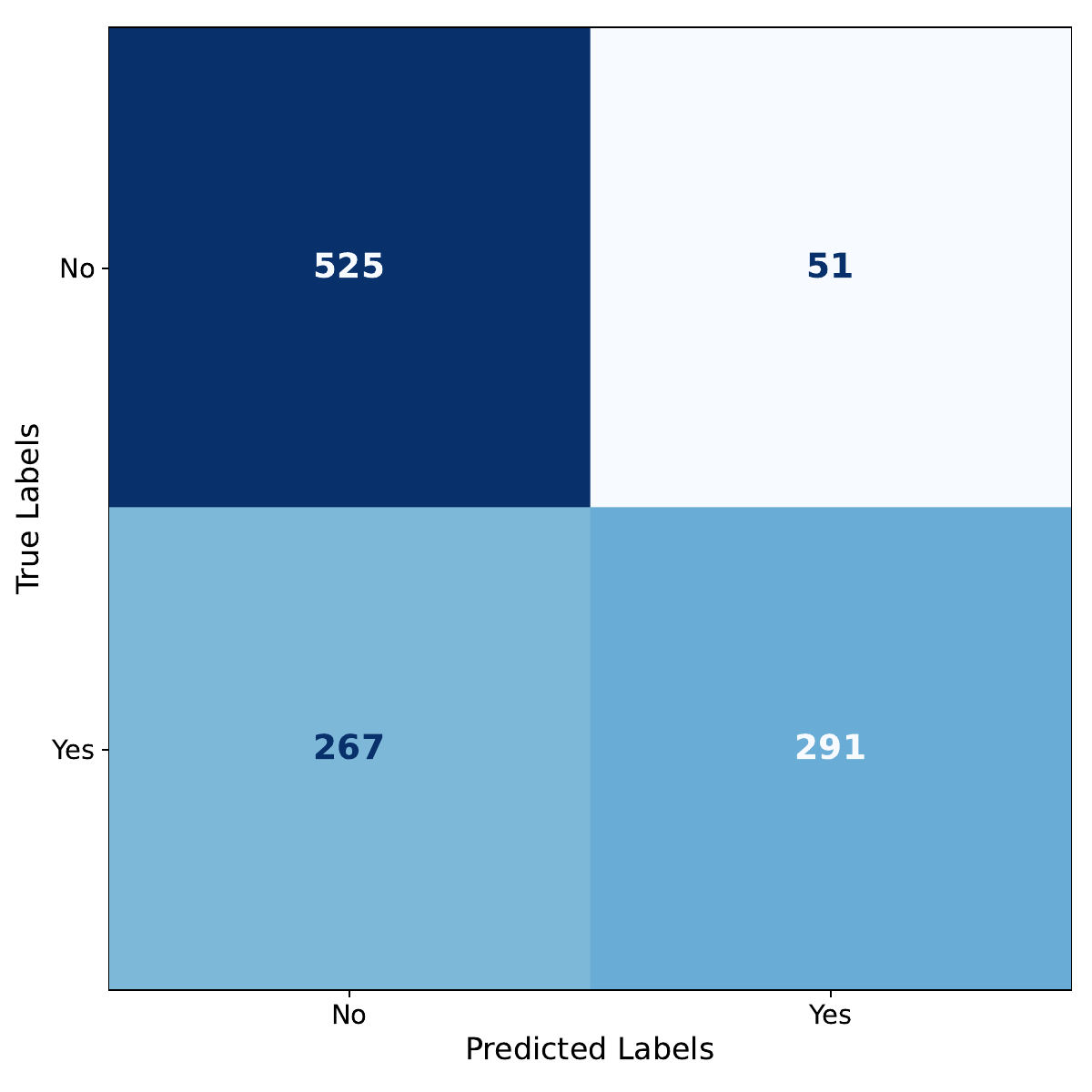}
         \caption{Qwen (Z+D)}
         \label{fig:qwen_ZD}
     \end{subfigure}
     \hfill
     \begin{subfigure}[b]{0.3\linewidth}
         \centering
         \includegraphics[width=\linewidth]{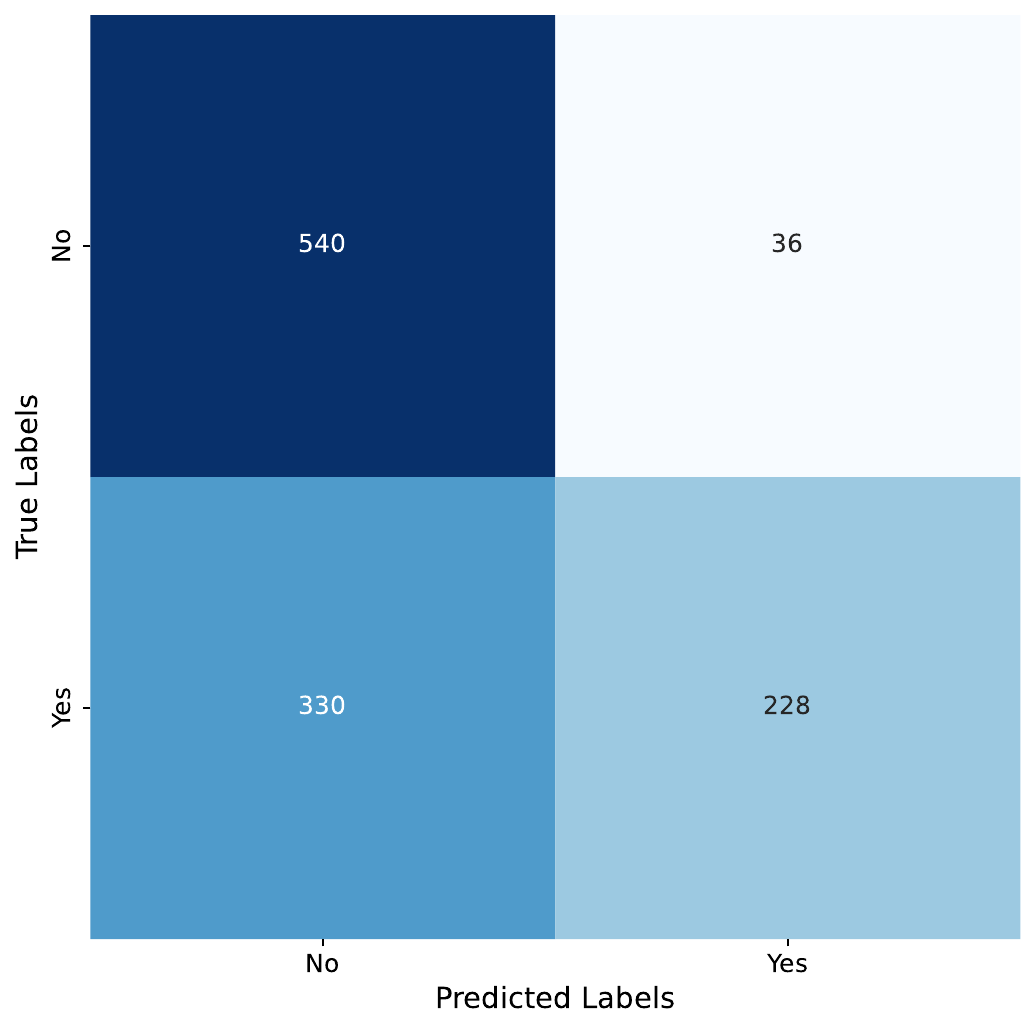}
         \caption{Qwen (F)}
         \label{fig:qwen_F}
     \end{subfigure}
        \caption{Confusion matrices for Qwen Z vs Z+D vs F.}
        \label{fig:confusion-qwen}
\end{figure*}

\begin{figure*}[t]
     \centering
     \begin{subfigure}[b]{0.3\linewidth}
         \centering
         \includegraphics[width=\linewidth]{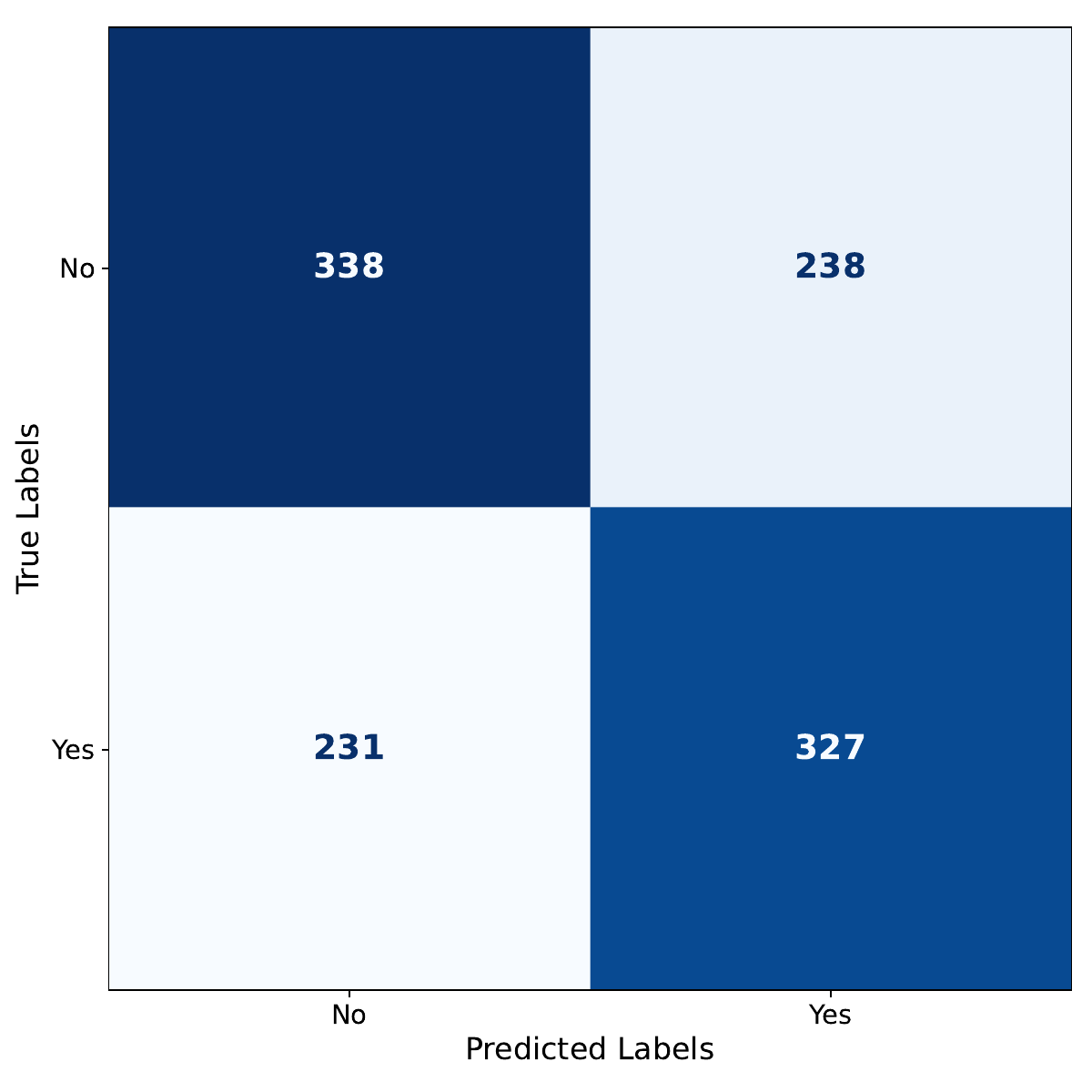}
         \caption{Gemma (Z)}
         \label{fig:gemma_Z}
     \end{subfigure}
     \hfill
     \begin{subfigure}[b]{0.3\linewidth}
         \centering
         \includegraphics[width=\linewidth]{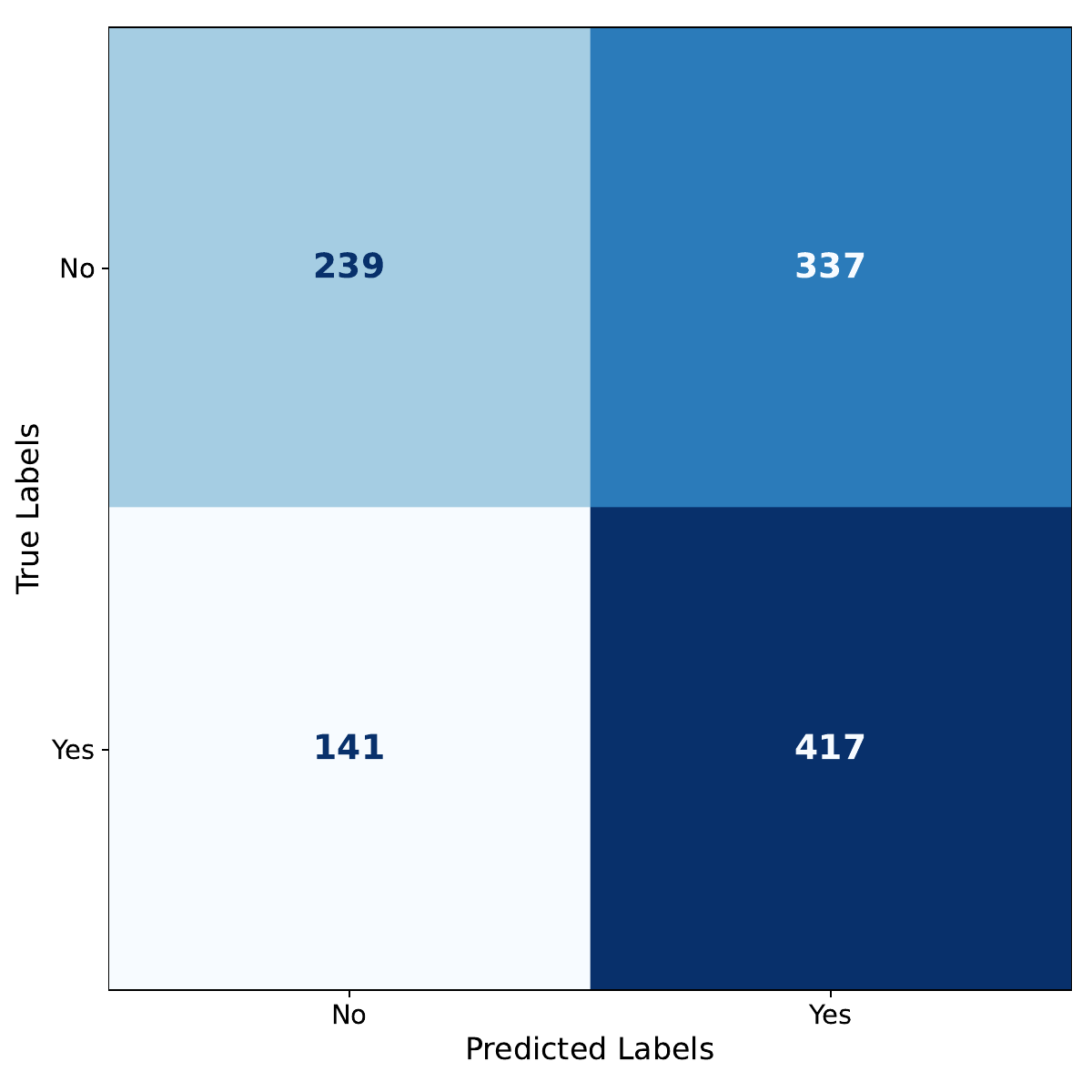}
         \caption{Gemma (Z+D)}
         \label{fig:gemma_ZD}
     \end{subfigure}
     \hfill
     \begin{subfigure}[b]{0.3\linewidth}
         \centering
         \includegraphics[width=\linewidth]{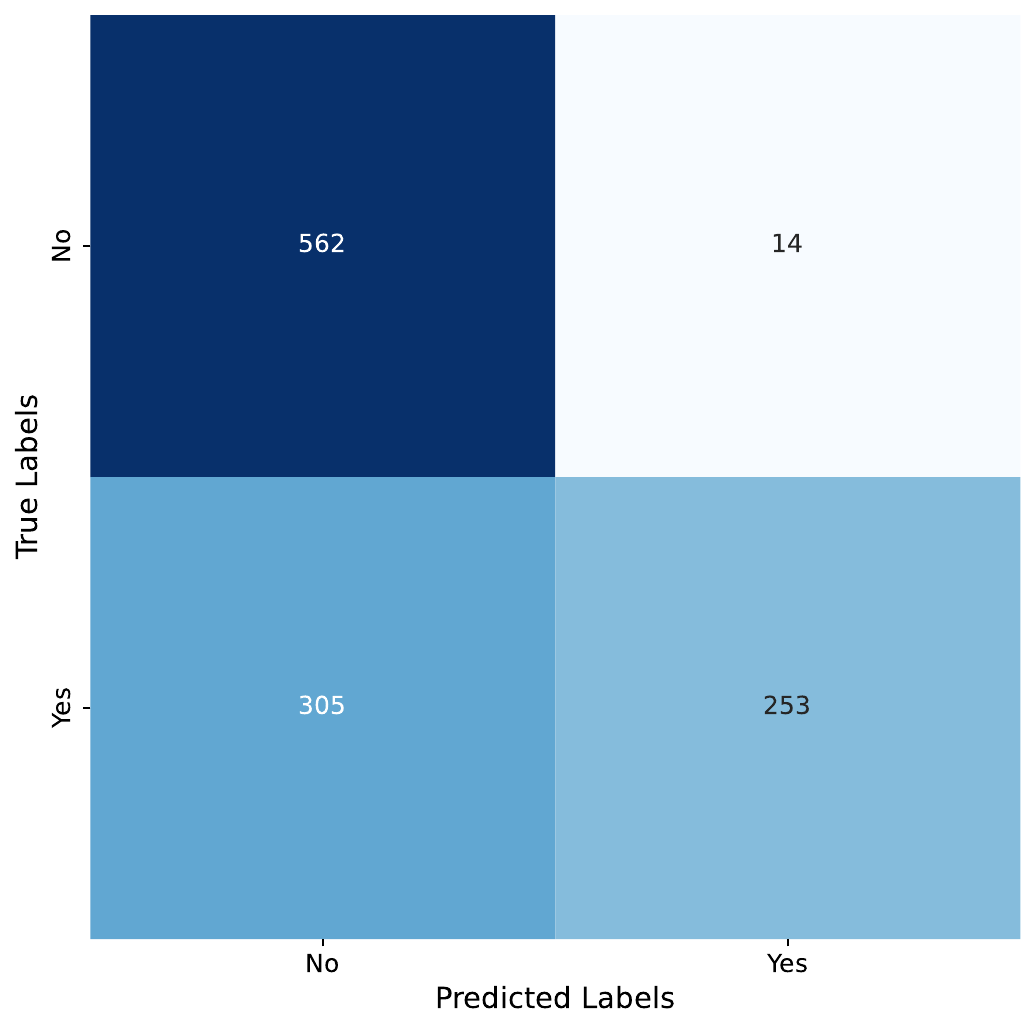}
         \caption{Gemma (F)}
         \label{fig:gemma_F}
     \end{subfigure}
        \caption{Confusion matrices for Gemma Z vs Z+D vs F.}
        \label{fig:confusion-gemma}
\end{figure*}

\begin{figure*}[t]
     \centering
     \begin{subfigure}[b]{0.45\linewidth}
         \centering
         \includegraphics[width=\linewidth]{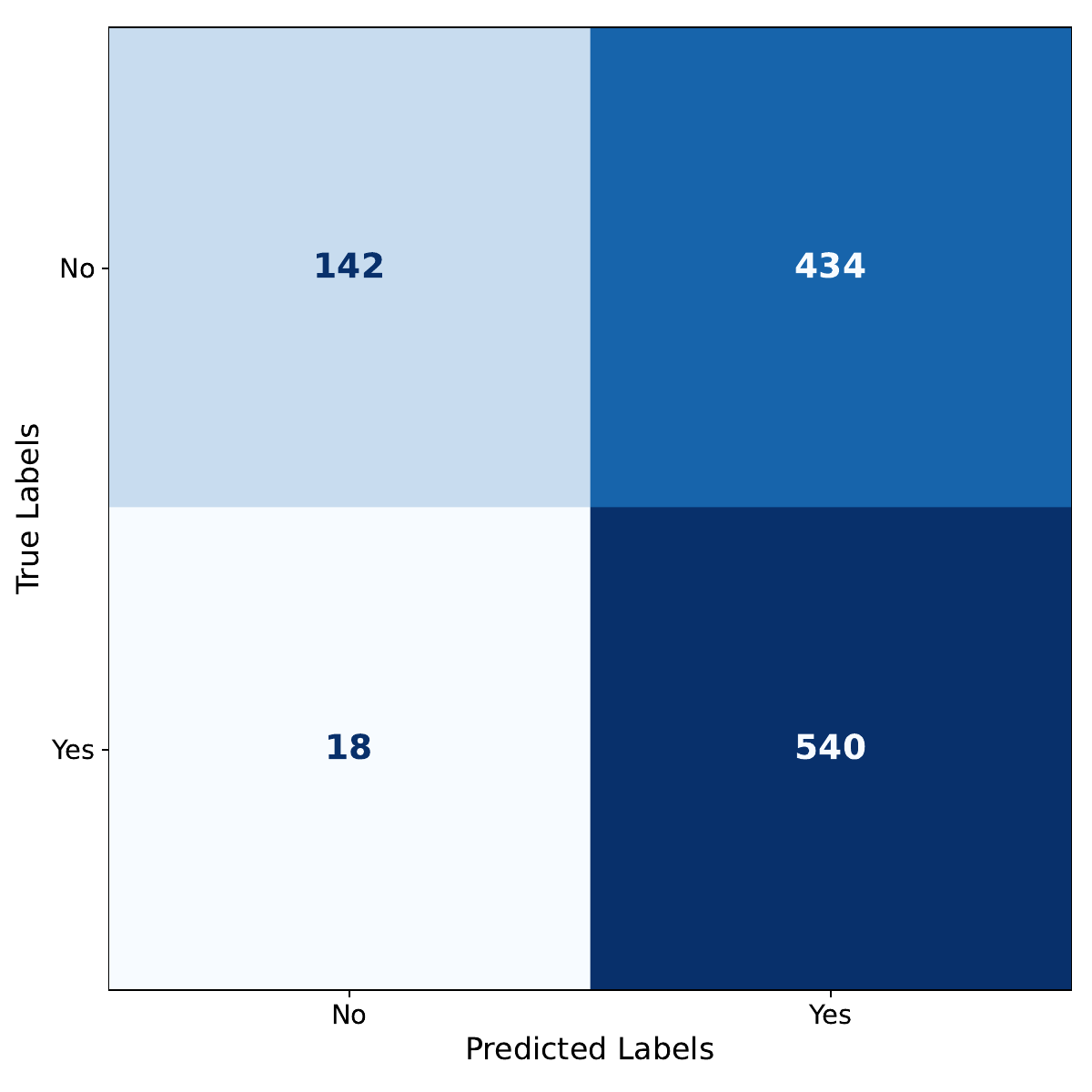}
         \caption{Mistral (Z)}
         \label{fig:mistral_Z}
     \end{subfigure}
     \hfill
     \begin{subfigure}[b]{0.45\linewidth}
         \centering
         \includegraphics[width=\linewidth]{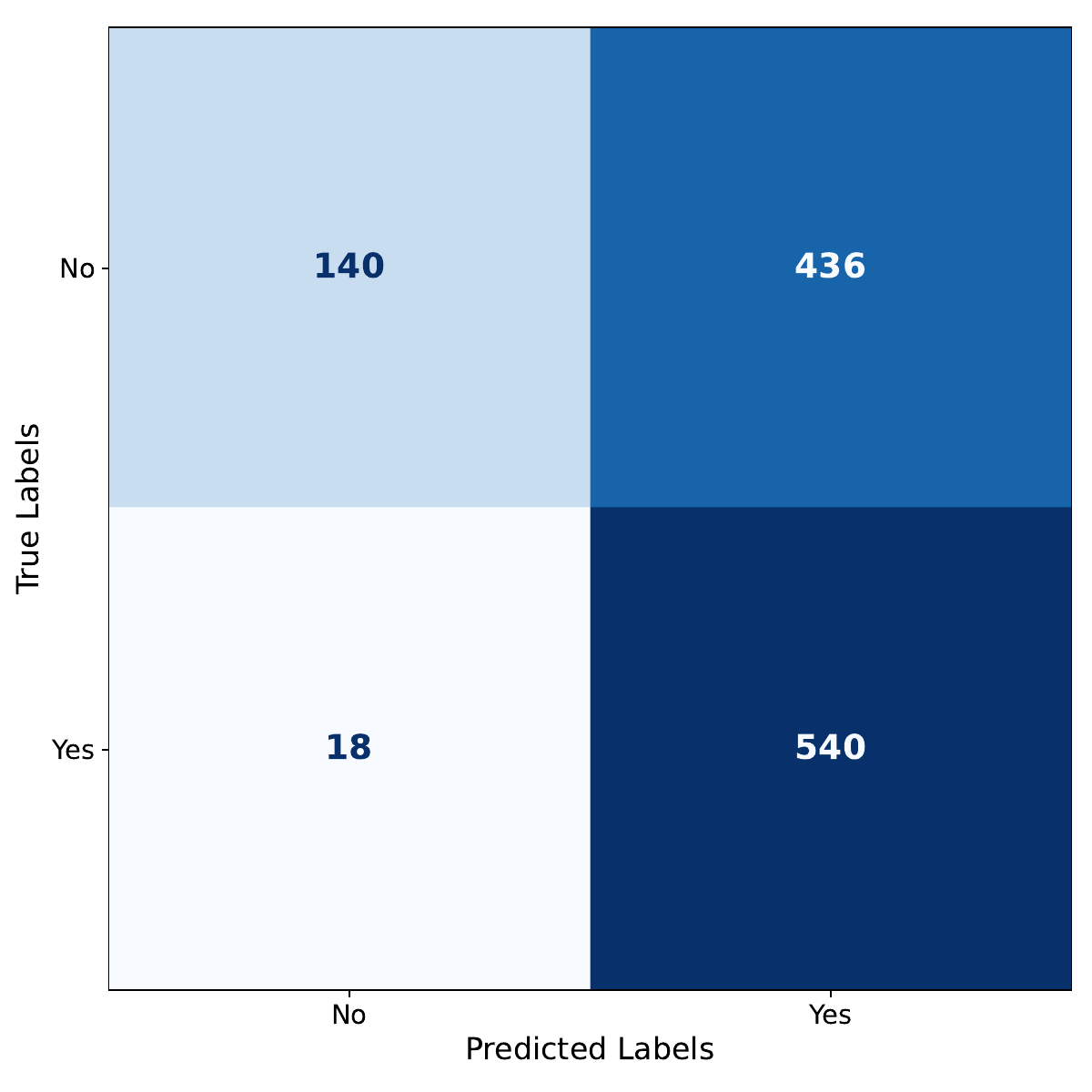}
         \caption{Mistral (Z+D)}
         \label{fig:mistral_ZD}
     \end{subfigure}
        \caption{Confusion matrices for Mistral Z vs Z+D.}
        \label{fig:confusion-mistral}
\end{figure*}

\begin{figure*}[t]
     \centering
     \begin{subfigure}[b]{0.45\linewidth}
         \centering
         \includegraphics[width=\linewidth]{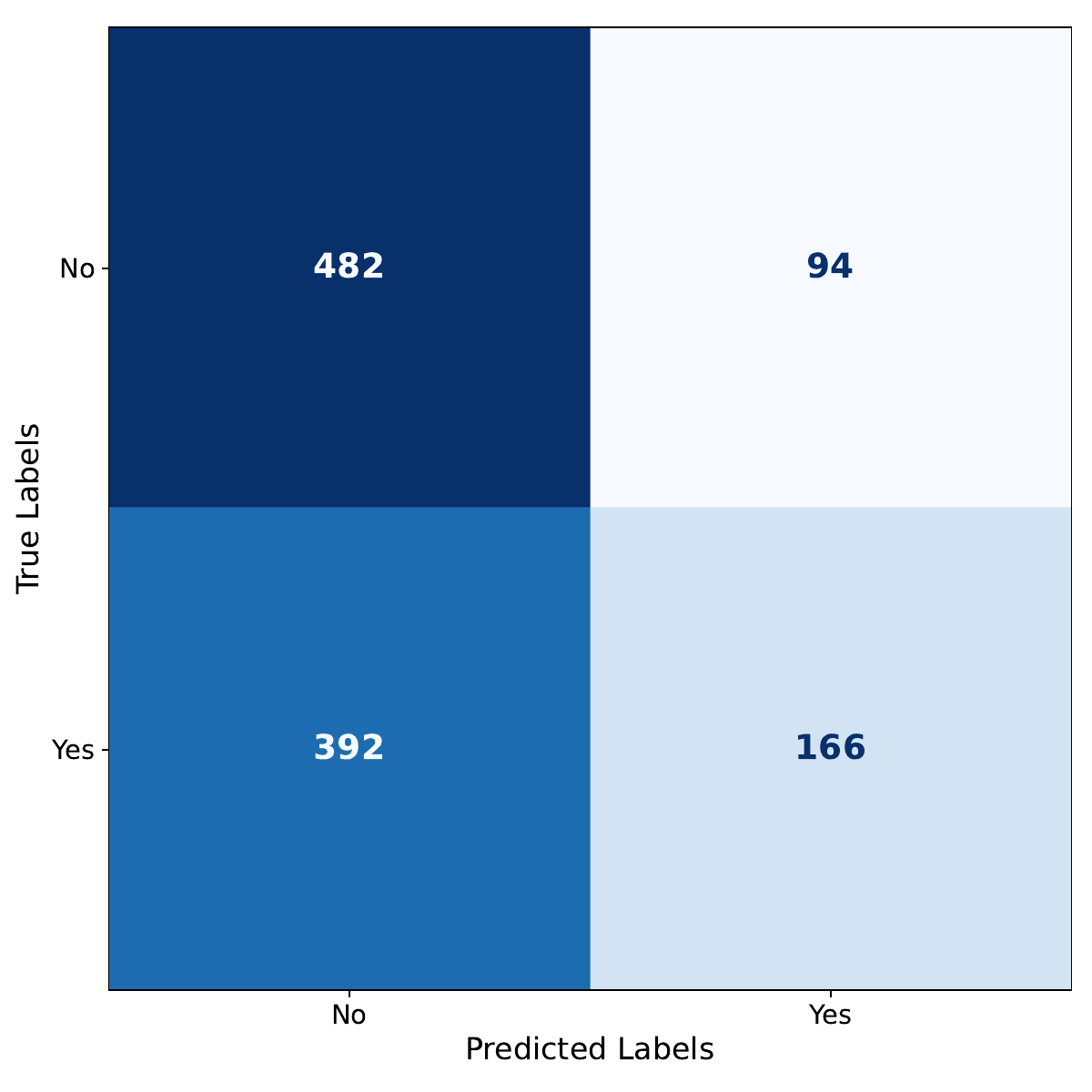}
         \caption{Vicuna (Z)}
         \label{fig:vicuna_Z}
     \end{subfigure}
     \hfill
     \begin{subfigure}[b]{0.45\linewidth}
         \centering
         \includegraphics[width=\linewidth]{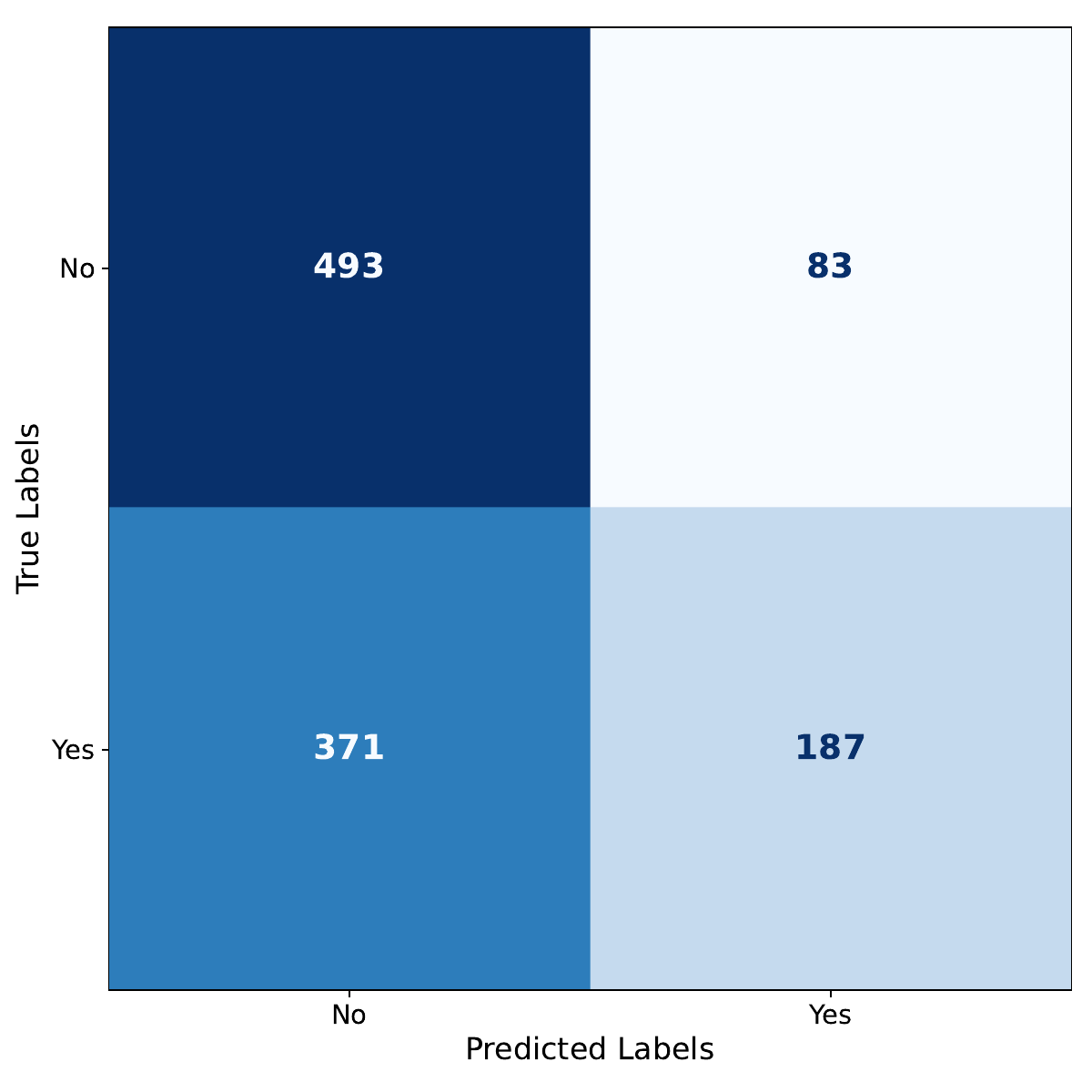}
         \caption{Vicuna (Z+D)}
         \label{fig:vicuna_ZD}
     \end{subfigure}
        \caption{Confusion matrices for Vicuna Z vs Z+D.}
        \label{fig:confusion-vicuna}
\end{figure*}

\begin{figure*}[t]
     \centering
     \begin{subfigure}[b]{0.45\linewidth}
         \centering
         \includegraphics[width=\linewidth]{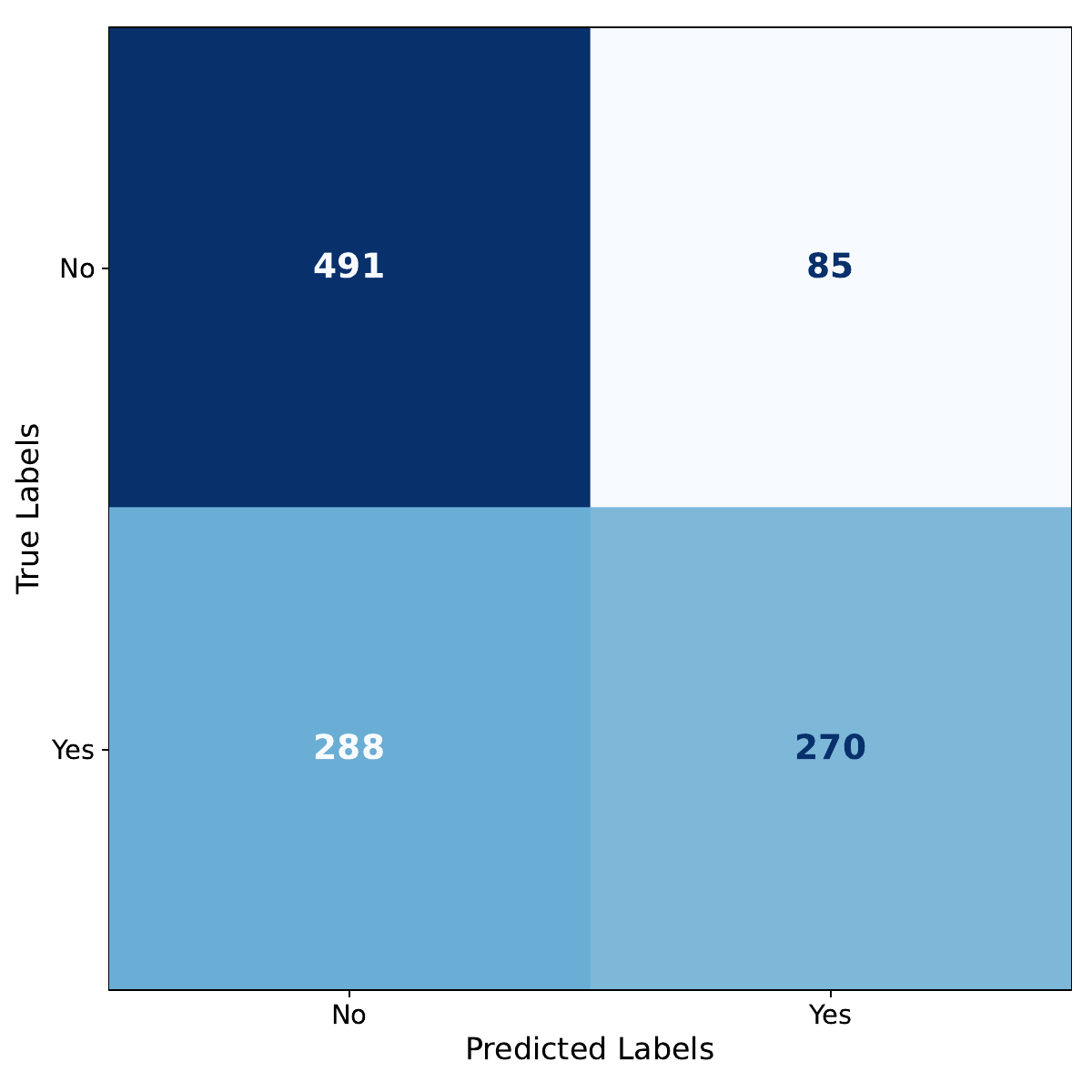}
         \caption{BERT (F)}
         \label{fig:bert_F}
     \end{subfigure}
     \hfill
     \begin{subfigure}[b]{0.45\linewidth}
         \centering
         \includegraphics[width=\linewidth]{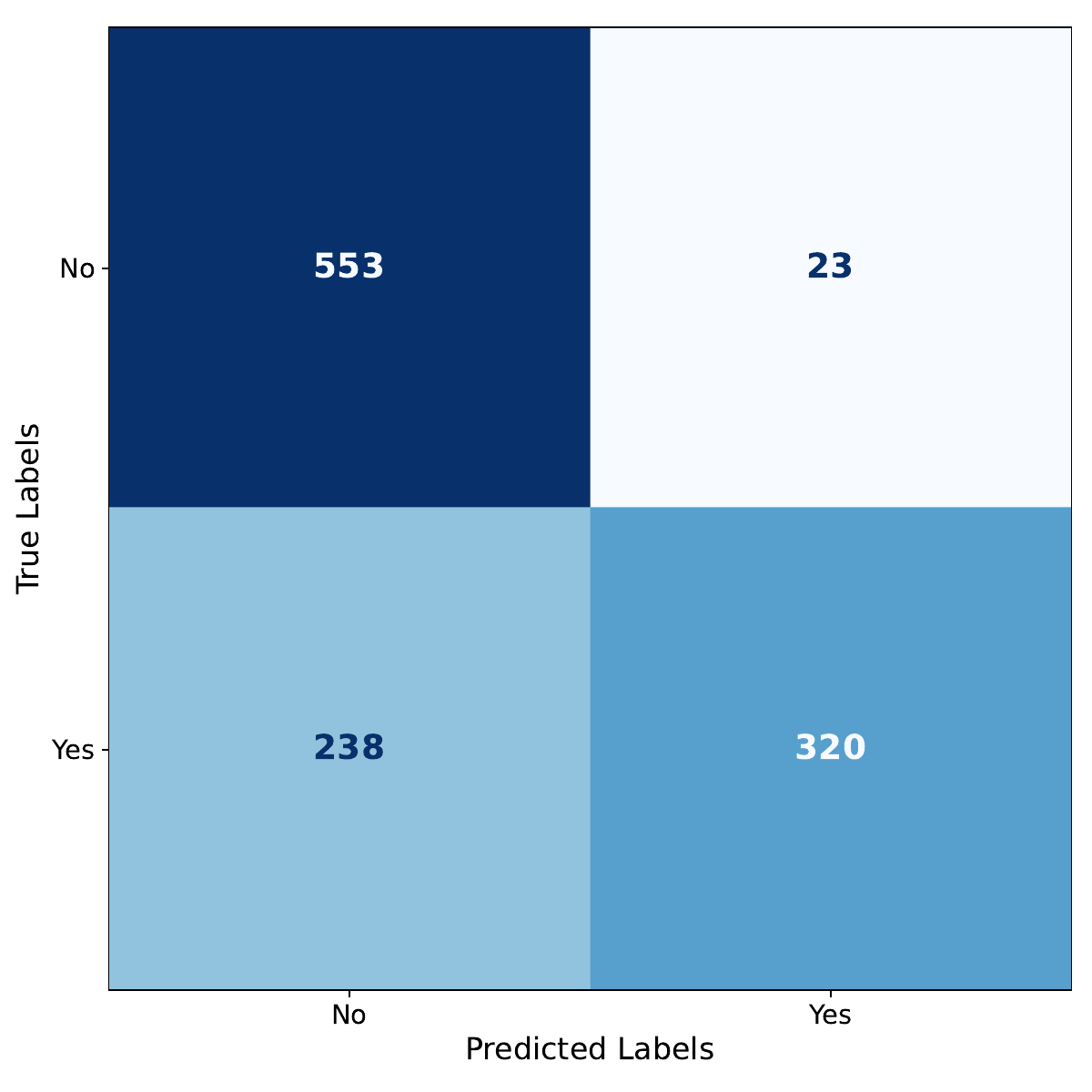}
         \caption{RoBERTa (F)}
         \label{fig:roberta_F}
     \end{subfigure}
        \caption{Confusion matrices for BERT (F) and RoBERTa (F).}
        \label{fig:confusion-encoders}
\end{figure*}

The confusion matrices for Qwen, Gemma, Mistral, Vicuna, BERT and RoBERTa are in \autoref{fig:confusion-qwen}, \autoref{fig:confusion-gemma}, \autoref{fig:confusion-mistral}, \autoref{fig:confusion-vicuna} and \autoref{fig:confusion-encoders} (BERT and RoBERTa) respectively.

\section{Performance Degradation Post Fine-Tuning}
\label{app:degradation}

We did a full fine-tuning on Vicuna but this had no effect on its performance. We suspect that Vicuna, being an older generation model, is not able to cope up with the newer models. This is also evident through Vicuna’s position towards the end of the leaderboards in  both Chatbot Arena (\url{https://lmarena.ai/?leaderboard}) as well as OpenLLM (\url{https://huggingface.co/spaces/open-llm-leaderboard/open_llm_leaderboard#/}).

We tried full fine-tuning on Mistral but due to resource constraints we could not explore it much. Our conjecture is that due to the small size of our dataset and the nuanced nature of our task, Mistral could have experienced catastrophic forgetting. This type of behavior is also reported by \citet{li-etal-2024-revisiting}.

For mitigation, future research directions could be increasing the dataset size, augmenting synthetic data with human-written data and exploring hyper-parameter tuning.

\section{Sentiment-Opposition Model Prompt}
\label{app:som}

\begin{lstlisting}
The Sentiment Opposition Model (SOM) for 
Humblebragging consists of the following 
components:

    1. Surface Sentiment (SS):
       - The apparent emotional tone of 
       the statement, usually negative 
       (complaint) 
       or neutral (modest).

    2. Intended Sentiment (IS):
       - The actual meaning the speaker 
       conveys, which is typically 
       positive and self-promotional.

    3. Sentiment Opposition (SO):
       - The contrast between SS and IS. 
       If SS is negative/neutral but IS 
       implies positive tone, 
       opposition exists.

    4. Humblebrag Classification:
       - If SO exists, classify the 
       statement as a humblebrag.
       - If SO doesn't exist, classify 
       as a non-humblebrag

Now you are about to classify if a given 
sentence is a humblebrag or not using 
the above definition.
\end{lstlisting}

\section{Humblebragging Component Identification Results}
\label{app:interpretation}

See \autoref{tab:combined_humblebrag} for component identification performance of three different models.

\begin{table*}[ht]
\centering
\resizebox{0.8\linewidth}{!}{%
\renewcommand{\arraystretch}{1.3} 
\setlength{\tabcolsep}{5pt} 
\begin{tabular}{p{2cm} p{14cm}} 
\toprule
\textbf{Example 1:} & \\ \midrule
\textbf{Input (Gold):} & 
{\color{blue}I can't believe} {\color{red}they'd give} {\color{blue}an idiot like me} {\color{red}a PhD} {\color{blue}lol} \quad (Brag Theme: Achievements; Mask Type: Complaint) \\[1ex]
\textbf{GPT-4o:} & 
{\color{blue}I can’t believe} {\color{red}they’d give} {\color{blue}an idiot like me} {\color{red}a PhD} {\color{blue}lol} \quad (Brag Theme: Achievement; Mask Type: Self-Deprecating) \\[1ex]
\textbf{Llama-3.1-8B:} & 
{\color{blue}I can’t believe} {\color{red}they’d give an idiot like me} {\color{blue}a PhD} \quad (Brag Theme: Achievements; Mask Type: Modesty) \\[1ex]
\textbf{Mistral-7B:} & 
{\color{blue}I can’t believe} {\color{red}they’d give an idiot like me} {\color{blue}a PhD} \quad (Brag Theme: Intelligence; Mask Type: Complaining) \\
\midrule

\textbf{Example 2:} & \\ \midrule
\textbf{Input (Gold):} & {\color{red} For the 3rd time in 3 years I’ve been asked to speak at Harvard,} {\color{blue} but I've yet to speak at my alma mater. What’s a girl gotta do @MarquetteU?} \quad (Brag Theme: Achievements; Mask Type: Complaint)\\[1ex]
\textbf{GPT-4o:} & {\color{red} For the 3rd time in 3 years I’ve been asked to speak at Harvard,} {\color{blue} but I've yet to speak at my alma mater. What’s a girl gotta do @MarquetteU?} \quad (Brag Theme: Achievements / Intelligence; Mask Type: Complaining) \\[1ex]
\textbf{Llama-3.1:} & {\color{red} For the 3rd time in 3 years I’ve been asked to speak at Harvard,} {\color{blue} but I've yet to speak at my alma mater.What’s a girl gotta do @MarquetteU?} \quad (Brag Theme: Achievements; Mask Type: Complaining) \\[1ex]
\textbf{Mistral-7B:} & {\color{red} For the 3rd time in 3 years I’ve been asked to speak at Harvard,} {\color{blue} I've yet to speak at my alma mater. What’s a girl gotta do @MarquetteU?} \quad (Brag Theme: Achievement; Mask Type: Complaining) \\ 
\midrule
\textbf{Example 3:} & \\ \midrule
\textbf{Input (Gold):} & {\color{red} Will Twitter be available for me in Paris, Milan, or the Maldives?} {\color{blue} I hope so bc it won't in Hong Kong or Singapore.} \quad (Brag Theme: Social Life; Mask Type: Complaint)\\[1ex]
\textbf{GPT-4o:} & {\color{red} Will Twitter be available for me in Paris, Milan, or the Maldives?} {\color{blue} I hope so bc it won't in Hong Kong or Singapore.} \quad (Brag Theme: Social Life / Wealth; Mask Type: Complaining) \\[1ex]
\textbf{Llama-3.1:} & {\color{blue} Will Twitter be available for me in Paris, Milan, or the Maldives?} {\color{red} I hope so bc it won't in Hong Kong or Singapore.} \quad (Brag Theme: Social Life; Mask Type: Modest) \\[1ex]
\textbf{Mistral-7B:} & {\color{red} Will Twitter be available for me in Paris, Milan, or the Maldives?} {\color{blue} I hope so bc it won't in Hong Kong or Singapore.} \quad (Brag Theme: Technology and Connectivity; Mask Type: Complaining) \\ 
\bottomrule
\end{tabular}
}
\caption{Humblebragging component identification: Model responses with identified brag (red), identified mask (blue), brag theme and mask type.}
\label{tab:combined_humblebrag}
\end{table*}

\end{document}